%% file: paperrevision1.tex
\documentclass[final,3p,times]{elsarticle}

\usepackage{graphicx}
\usepackage{amssymb}

\usepackage[ruled,vlined]{algorithm2e}
\usepackage{lineno}
\input headerDD

\usepackage{times,cite,color,soul}
\usepackage{amsmath}

\begin{document}
\pagenumbering{arabic}
\begin{frontmatter}


\title{Sample-to-Sample Correspondence for Unsupervised Domain Adaptation}




\author{Debasmit Das}
\ead{debasmit.das@gmail.com}
\author{C.S. George Lee}
\ead{csglee@purdue.edu}
\address{School of Electrical and Computer Engineering, Purdue University, West Lafayette, Indiana, USA}

\begin{abstract}
The assumption that training and testing samples 
are generated from the same distribution does not always hold 
for real-world machine-learning applications. 
The procedure of tackling this discrepancy between the training (source) 
and testing (target) domains is known as domain adaptation. 
We propose an unsupervised version of domain adaptation 
that considers the presence of only unlabelled data in the target domain. 
Our approach centers on finding correspondences between samples of each domain.
The correspondences are obtained by treating the source and target samples 
as graphs and using a convex criterion to match them. 
The criteria used are first-order and second-order similarities 
between the graphs as well as a class-based regularization. We have also developed a computationally efficient routine 
for the convex optimization, thus allowing the proposed method to be used widely. 
To verify the effectiveness of the proposed method, 
computer simulations were conducted on synthetic, image classification 
and sentiment classification datasets. 
Results validated that the proposed local sample-to-sample matching method 
out-performs traditional moment-matching methods 
and is competitive with respect to current local domain-adaptation methods. 
\end{abstract}


\end{frontmatter}


\section{Introduction}
\label{S:1}

\begin{figure}[h]
\centering
\includegraphics[width=8cm]{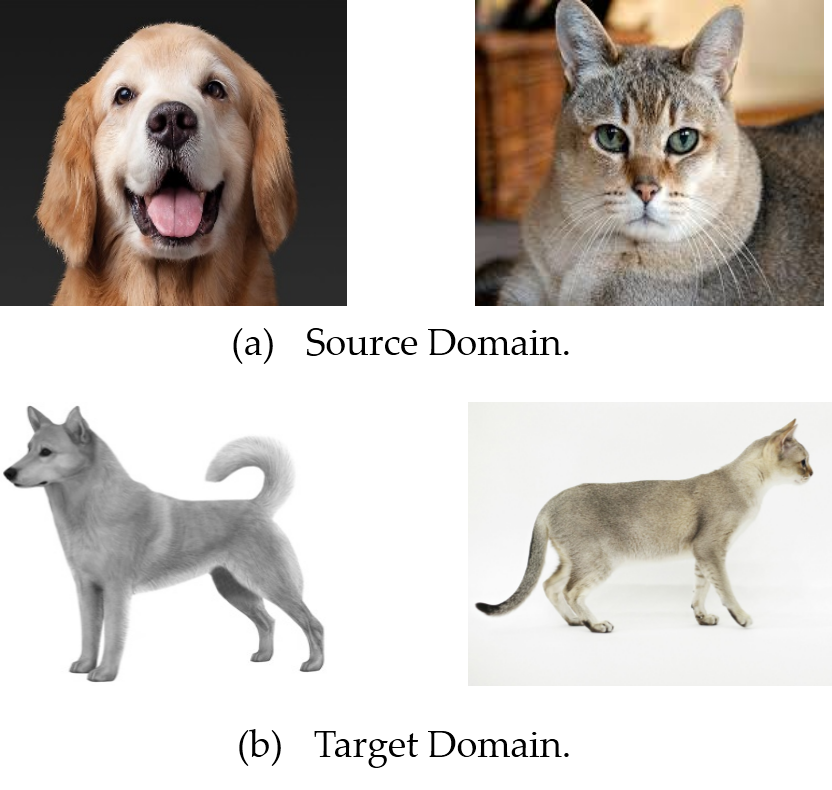}
\vspace*{-0.1in}
\caption{Discrepancy between the source domain and the target domain. 
In the source domain, the images have frontal faces 
while the target domain has images of the whole body from the side view-point.}
\label{cat-dog}
\vspace*{-0.1in}
\end{figure}

In traditional machine-learning settings, 
we assume that the testing data belongs to the same distribution as the training data. 
However, such an assumption is rarely encountered in real-world situations. 
For example, consider a recognition system that distinguishes 
between a cat and a dog, given labelled training samples of the type 
shown in Fig.~\ref{cat-dog}(a). 
These training samples are frontal faces of cats and dogs. 
When the same recognition system is used to test in a different domain 
such as on the side images of cats and dogs as shown in Fig.~\ref{cat-dog}(b), 
it would fail miserably. 
This is because the recognition system has developed 
a bias in being able to only distinguish between the face of a dog and a cat 
and not side images of dogs and cats. 
Domain adaptation (DA) aims to mitigate this dataset bias~\citep{torralba2011unbiased}, where different datasets have their own unique properties.
Dataset bias appears because of the distribution shift of data 
from one dataset (i.e., source domain) to
another dataset (i.e., target domain). 
The distribution shift manifests itself in different forms. 
In computer vision, it can occur when there is changing lighting conditions, 
changing poses, etc. 
In speech processing, it can be due to changing accent, 
tone and gender of the person speaking. 
In remote sensing, it can be due to changing atmospheric conditions, 
change in acquisition devices, etc. 
To encounter this discrepancy in distributions, 
domain adaptation methods have been proposed. 
Once domain adaptation is carried out, 
a model trained using the adapted source domain data 
should perform well in the target domain. 
The underlying assumption in domain adaptation 
is that the task is the same in both domains. 
For classification problems, it implies that we have the same set of categories
in both source and target domains. 

Domain adaptation can also assist in annotating datasets efficiently 
and further accelerating machine-learning research. 
Current machine-learning models are data hungry 
and require lots of labelled samples. 
Though huge amount of unlabelled data is obtained, 
labelling them requires lot of human involvement and effort. 
Domain adaptation seeks to automatically annotate unlabelled data 
in the target domain by adapting the labelled data 
in the source domain to be close to the unlabeled target-domain data. 

In our work, we consider \emph{unsupervised domain adaptation} (UDA), 
which assumes absence of labels in the target domain. 
This is more realistic than semi-supervised domain adaptation, 
where there are also a few-labelled data in the target domain. 
This is because labelling data might be time-consuming 
and expensive for real-world situations. 
Hence we need to effectively exploit fully labelled source-domain data 
and fully unlabelled target-domain data to carry out domain adaptation. 
In our case, we seek to find correspondences 
between each source-domain sample and each target-domain sample. 
Once the correspondences are found, 
we can transform the source-domain samples 
to be close to the target-domain samples. 
The transformed source-domain samples will then lie close to 
the data space of the target domain. 
This will allow a model trained on the transformed source-domain data 
to perform well with the target-domain data. 
This not only achieves the goal of training robust models 
but also allows the model to annotate unlabelled target-domain data accurately.

The remainder of the paper is organized as follows: 
Section 2 discusses related work of domain adaptation. 
Section 3 discusses the background required for our proposed approach. 
Section 4 discusses our proposed approach 
and formulates our unsupervised domain adaptation problem into 
a constrained convex optimization problem. 
Section 5 discusses the experimental results and some comparison with existing work. Section 6 discusses some limitations. 
Section 7 concludes with a summary of our work and future research directions.
Finally, the Appendix shows more details about 
the proof of convexity of the optimization objective function and derivation of the gradients.

\section{Related Work}
There is a large body of prior work on domain adaptation. 
For our case, we only consider homogeneous domain adaptation, 
where both the source and target domains have the same feature space. 
Most of previous DA methods are classified into two categories, 
depending on whether a deep representation is learned or not. 
In that regard, our proposed approach is not deep-learning-based since 
we directly work at the feature level without learning a representation. 
We feel that our method can easily be extended to deep architectures 
and provide much better results. 
For a comprehensive overview on domain adaptation, 
please refer to Csurka's survey paper~\citep{csurka2017domain}.

\subsection{Non-Deep-Learning Domain-Adaptation Methods}
These non-deep-learning domain-adaptation methods can be broadly 
classified into three categories -- instance re-weighting methods, 
parameter adaptation methods, and feature transfer methods. 
Parameter adaptation methods~\citep{jiang2008cross,bruzzone2010domain,
duan2009domain,yang2007cross} generally adapt a trained classifier 
in the source domain (e.g., an SVM) in order 
to perform better in the target domain. 
Since these methods require at least a small set of labelled target examples, 
they cannot be applied to UDA.

\emph{Instance Re-weighting} was one of the early methods, 
where it was assumed that conditional distributions 
were shared between the two domains. 
The instance re-weighting involved estimating 
the ratio between the likelihoods of being a source example 
or a target example to compute the weight of an instance. 
This was done by estimating the likelihoods independently~\citep{zadrozny2004learning} 
or by approximating the ratio between the densities~\citep{kanamori2009efficient,sugiyama2008direct}. 
One of the most popular measures used to weigh data instances, 
used in~\citep{gretton2009covariate,huang2007correcting}, 
was the Maximum Mean Discrepancy (MMD)~\citep{borgwardt2006integrating} 
calculated between the distributions in different domains. 
\emph{Feature Transfer} methods, on the other hand, 
do not assume the same conditional distributions 
between the source and target domains. 
An early method for Domain Adaptation was proposed in~\citep{daume2009frustratingly}, 
where the representation is modifed such that
the source features are $(\bx_s,\bx_s,0)$ 
and the target features are $(\bx_t,0,\bx_t)$. 
This enables identifying shared and domain-specific features. 
Other ideas include the Geodesic Flow Sampling (GFS)~\citep{gopalan2014unsupervised,gopalan2011domain} 
and the Geodesic Flow Kernel (GFK)~\citep{gong2012geodesic,gong2013connecting}, 
where the domains are considered as samples on the Grassman manifolds. The Subspace Alignment (SA)~\citep{fernando2013unsupervised} method
aligns source and target domain subspaces using Bregman divergence. 
The linear Correlation Alignment (CORAL)~\citep{sun2016return} 
algorithm aligns the source and target data covariances. 
Transfer Component Analysis (TCA)~\citep{pan2011domain} 
discovers shared hidden  features having simmilar distribution 
between the two domains. 
Chen at al.\citep{chen2012marginalized} proposed a reconstruction based approach to learn a domain invariant representation.
Most of these previous methods learned global alignment 
between the two domains. 
On the other hand, the Adaptive Transductive Transfer Machines (ATTM)~\citep{farajidavar2014adaptive} and Optimal Transport \citep{courty2016optimal} considers sample to sample alignment between the source and target distributions. 

\subsection{Deep Domain-Adaptation Methods}
Most deep-learning methods for DA use
a siamese architecture with two streams for the source and target domain. 
These methods use classification loss in addition to a discrepancy 
loss~\citep{long2016deep,long2015learning,tzeng2014deep,ghifary2015domain,deepCoral} 
or an adversarial loss. 
The classification loss depends on the labelled source data, 
and the discrepancy loss diminishes the shift between the two domains. 
On the other hand, adversarial-based methods
play a game of generating domain-invariant representations with the domain discriminator. 
\citet{tzeng2017adversarial} proposes a unified view 
of existing adversarial DA methods by comparing them according to the loss type, 
the weight-sharing strategy between the two streams, and on whether they are discriminative or generative. 
The Domain-Adversarial Neural Networks (DANN)~\citep{ganin2016domain} 
used a gradient reversal layer to produce features that are discriminative as well as domain-invariant. 
The main disadvantage of these adversarial methods is that their training is generally not stable. 
Moreover, empirically tuning the capacity of a discriminator requires lot of effort.

Between these two classes of DA methods, 
the state-of-the-art methods are dominated by deep architectures. 
However, these approaches are quite complex and expensive, 
requiring re-training of the network and tuning of many hyper parameters 
such as the structure of the hidden adaptation layers. 
Non-deep-learning domain-adaptation methods do not achieve as good performance 
as a deep-representation approach, 
but they work directly with shallow/deep features 
and require lesser number of hyper-parameters to tune.  
Among the non-deep-learning domain-adaptation methods, 
we feel feature transformation methods are more generic 
because they directly use the feature space from the source and target domains, 
without any underlying assumption of the classification model. 
In fact, a powerful shallow-feature transformation method 
can be extended to deep-architecture methods, if desired, 
by using the features of each and every layer and then jointly optimizing 
the parameters of the deep architectures as well as that of the classification model. 
For example, correlation alignment~\citep{sun2016return} 
has been extended for deep architectures~\citep{deepCoral}, 
which evidently achieve the state-of-the art performance. 
Moreover, we believe a local transformation-based approach 
as in~\citep{courty2016optimal,farajidavar2014adaptive} 
will result in better performance than global transformation methods 
because it considers the effect of each and every sample in the dataset explicitly. 

\section{Background}
\begin{figure*}[htb]
\centering
\includegraphics[width=14cm]{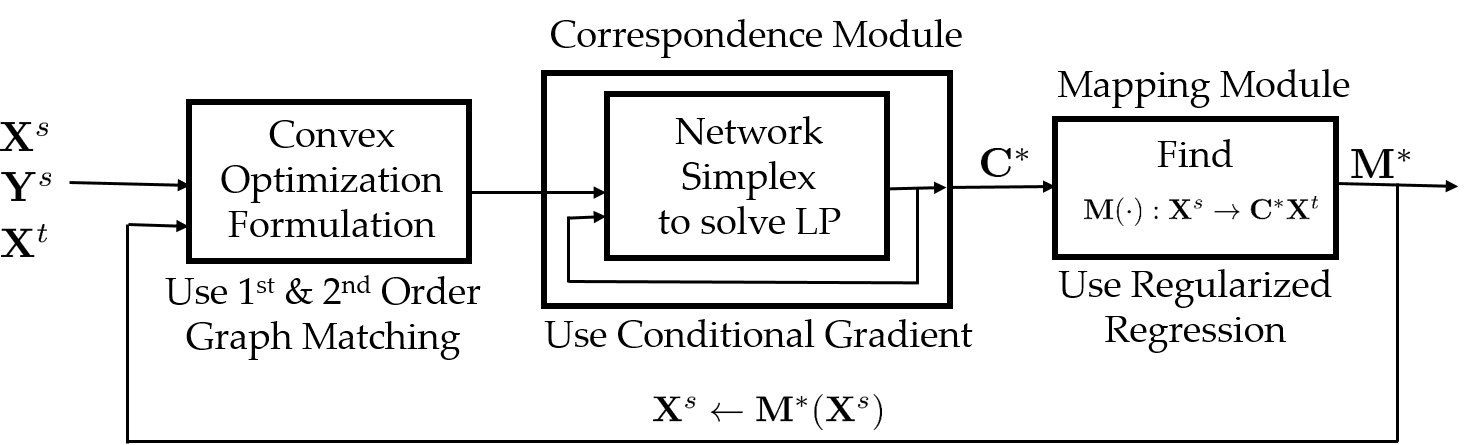}
\caption{Conceptual and high-level description of our proposed convex optimization formulation with its proposed solution. 
The inputs are source-domain data ($\bX^{s}$), source-domain labels ($\bY^{s}$), 
and target-domain data ($\bX^{t}$). 
Output is a mapping function ($\bM(\cdot)$) that maps $\bX^{s}$ close to $\bX^{t}$. 
The transformation can be repeated again by providing the transformed source data 
$\bM(\bX^{s})$, source labels $\bY^{s}$ and target data $\bX^{t}$ as input.}
\label{whole}
\end{figure*}

Our local transformation-based approach 
to DA places a strong emphasis on establishing a sample-to-sample
correspondence between each source-domain sample 
and each target-domain sample. 
Establishing correspondences between two sets of visual features 
have long been used in computer vision mostly for
image registration~\citep{besl1992method,chui2003new}. 
To our knowledge, the approach of finding correspondences 
between the source-domain and the target-domain samples 
has never been used for domain adaptation. 
The only work that is similar to finding correspondences 
is the work on optimal transport~\citep{courty2016optimal}. 
They learned a transport plan for each source-domain sample 
so that they are close to the target-domain samples. 
Their transport plan is defined on a point-wise unary cost 
between each source sample and each target sample. 
Our approach develops a framework to find correspondences between the source 
and target domains that exploit higher-order relations
beyond these unary relations between the source and target domains. 
We treat the source-domain data and
the target-domain data as the source and target hyper-graphs, respectively, 
and our correspondence problem
can be cast as a hyper-graph matching problem. 
The hyper-graph matching problem has been previously 
used in computer vision~\citep{duchenne2011tensor} through a tensor-based
formulation but has not been applied to domain adaptation. 
Hyper-graph matching involves using higher-order relations between samples 
such as unary, pairwise, tertiary or more. 
Pairwise matching involves matching source-domain sample pairs 
with target-domain sample pairs. 
Tertiary matching involves matching source-domain sample triplets 
with target-domain sample triplets and so on. 
Thus, hyper-graph methods provide additional higher-order geometric 
and structural information about the data that is missing
with just using unary point-wise relations between a source sample and a target sample. 
\begin{figure}[h]
\centering
\includegraphics[width=8cm]{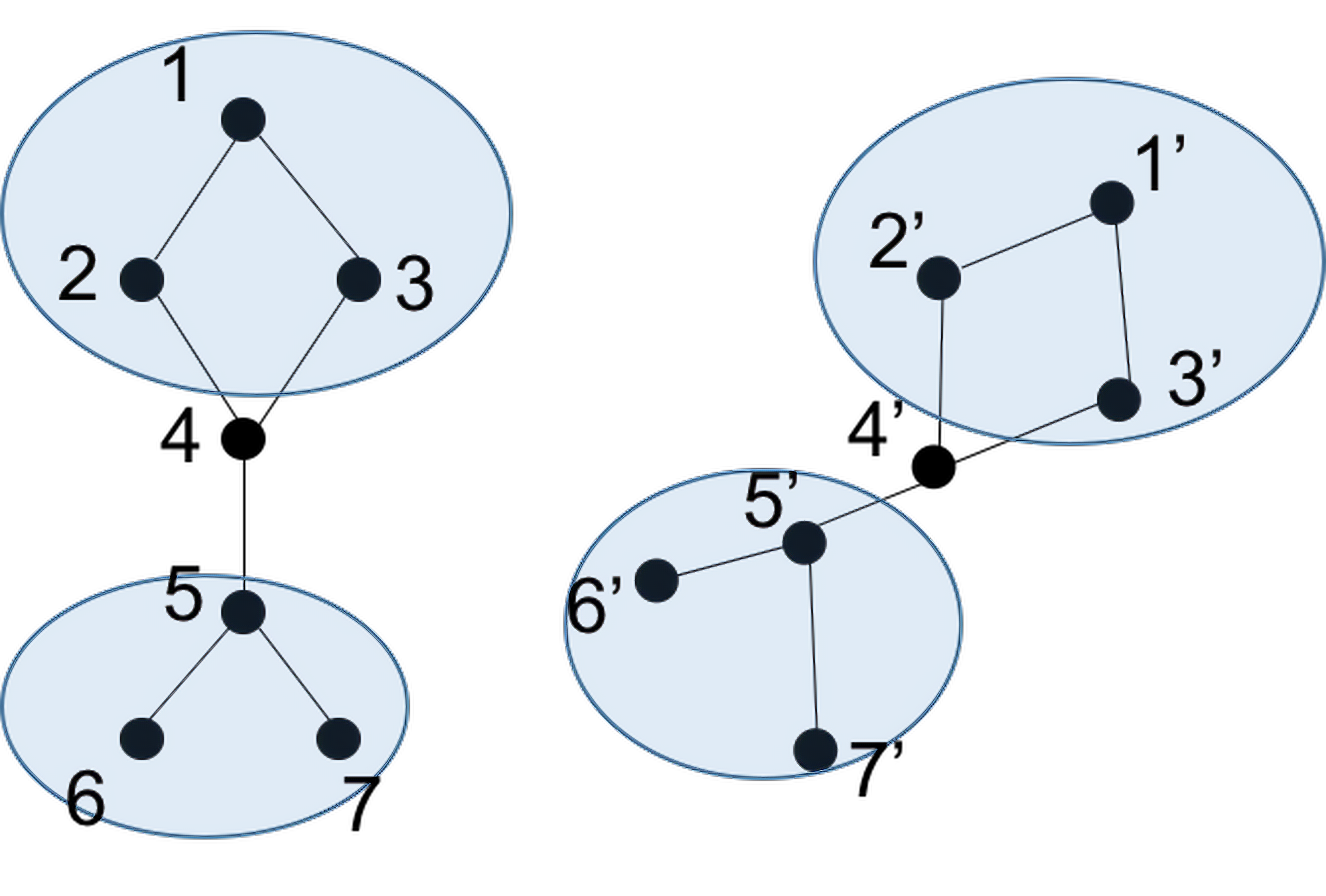}
\caption{Example showing the advantage of higher-order graph matching compared to just first-order matching.}
\label{HO}
\end{figure}
The advantage of using higher-order information in graph matching 
is demonstrated in the example in Fig.~\ref{HO}. 
In Fig.~\ref{HO}, the graph on the left is constructed 
from the source domain while the graph on the right is constructed from the target domain. 
In the graph, each node represents a sample and edges represent connectivity among the samples. 
Among these, samples $1$ and $1'$ do not match because those samples are not the closest pair of samples. 
But as a group $\{1,2,3\}$ matches with $\{1',2',3'\}$ suggesting that higher-order matching 
can aid domain adaptation, whereas one-to-one matchings 
between samples might not provide enough or provide incorrect information. 
Unfortunately, higher-order graph matching comes with
increasing computational complexity 
and also extra hyper-parameters that weigh the importance of each of the higher-order relations. 
Therefore , in our work we consider only the first-order 
and second-order matchings to validate the approach. 
Still, our problem can be inefficient because the number of correspondence variables 
increases with the number of samples. 
To address all these problems, we contribute in the following ways:
\begin{enumerate}
\item 
We initially propose a mathematical framework 
that uses the first-order and second-order relations 
to match the source-domain data and the target-domain data. 
Once the relations are established, 
the source domain is mapped to be close to the target domain. 
A class-based regularization is also used to leverage the labels present in the source domain. 
All these cost factors are combined into a convex optimization framework. 
\item 
The above transformation approach is computationally inefficient. 
We then reformulate our convex optimization problem 
into solving a series of sub-problems for which an efficient solution using a network simplex approach. 
This new formulation is more efficient in terms of both time and storage space.
\item 
Finally, we have performed experimental evaluation of our proposed method on both toy datasets 
as well as real image and sentiment classification datasets. 
We have also examined the effect of each cost term in the convex optimization problem separately. 
\end{enumerate}
The overall scheme of our proposed approach is shown in Fig.~\ref{whole}
\section{Proposed Sample-to-Sample Correspondence
Method}
In this section,
we shall first define the domain adaptation problem~\citep{survey1,survey2},
and then formulate the proposed correspondence-and-mapping method
for the unsupervised domain adaptation problem.

\vspace*{-0.1in}
\subsection{Notation}
A domain consists of a $d$-dimensional 
feature space $\mathcal{X} \subset \mathbb{R}^d$ with a marginal probability 
distribution $P(\bX)$. The task $\mathcal{T}$ is defined using a label space
$\mathcal{Y}$ and the conditional probability distribution $P(\bY|\bX)$. Here $\bX$ and $\bY$ are random variables. 
For a particular sample set $\bX=\{\bx_1, \ldots,\bx_n\}$ of  $\mathcal{X}$
with labels $\bY=\{{y}_1, \ldots, {y}_n\}$ 
from $\mathcal{Y}$, $P(\bY|\bX)$ can be trained in a supervised way from feature and labels $\{\bx_i, y_i\}$. 
For the domain adaptation purpose, 
we assume that there are two domains with the same task: 
a source domain $\mathcal{D}^s=\{\mathcal{X}^s, P(\bX^s)\}$ 
with $\mathcal{T}^s=\{\mathcal{Y}^s, P(\bY^s|\bX^s)\}$ and 
a target domain $\mathcal{D}^t=\{\mathcal{X}^t, P(\bX^t)\}$ 
with $\mathcal{T}^t=\{\mathcal{Y}^t, P(\bY^t|\bX^t)\}$. 
Traditional machine learning techniques assume 
that both $\mathcal{D}^s=\mathcal{D}^t$ and $\mathcal{T}^s=\mathcal{T}^t$,  
where $\mathcal{D}^s$ becomes the training set and $\mathcal{D}^t$ the test set.
For domain adaptation, $\mathcal{D}^t \neq \mathcal{D}^s$ but $\mathcal{T}^t = \mathcal{T}^s$. 
When the source domain is related to the target domain, 
it is possible to use the relational information 
from ${\mathcal{D}^s,\mathcal{T}^s}$ to learn $P(\bY^t|\bX^t)$. 
The presence/absence of labels in the target domain also decide 
how domain adaptation is being carried out. 
We shall solve the most challenging case, 
where we have labelled source domain data but unlabeled data in the target domain. 
This is commonly known as unsupervised domain adaptation (UDA). 
A natural extension to UDA is the semi-supervised case, 
where a small set of target domain samples is labelled.

In our case, we have labelled source-domain data with 
a set of training data $\bX^s = \{\bx^s_i\}^{n_s}_{i=1}$ 
associated with a set of class labels $\bY^s = \{y^s_i\}^{n_s}_{i=1}$. 
In the target domain, we only have unlabelled samples $\bX^t = \{\bx^t_i\}^{n_t}_{i=1}$. 
If we had already trained a classifier using the source-domain samples, 
the performance of the target-domain samples on that classifier would be quite poor. 
This is because the distributions of the source and target samples are different; 
that is, $P(\bX^s) \neq P(\bX^t)$. 
So we need to to find a transformation of the input space 
$\bF : \mathcal{X}^s \rightarrow \mathcal{X}^t$ such that $P(y|\bx^t) = P(y|\bF(\bx^s))$. 
As a result of this transformation, 
the classifier learned on the transformed source samples 
can perform satisfactorily on the target-domain samples.

\subsection{Correspondence-and-Mapping Problem Formulation}
With the above notation,
our proposed approach considers the transformation $\bF$ 
as a point-set registration between two point sets, 
where the source samples $\{\bx^s_i\}^{n_s}_{i=1}$ are the moving point set 
and the target samples $\{\bx^t_i\}^{n_t}_{i=1}$ are the fixed point set. 
In such a case, the registration involves alternately finding the correspondence and mapping 
between the fixed and moving point sets~\citep{chui2003new,besl1992method}. 
The advantage of point-set registration is that it ensures explicit sample-to-sample matching 
and not moment matching like covariance in CORAL~\citep{sun2016return} 
or MMD~\citep{long2016deep,long2015learning,tzeng2014deep,ghifary2015domain}. 
As a result, the transformed source domain matches better with the target domain. 
However, matching each and every sample requires an optimizing variable 
for each pair of source and target domain samples. 
If the number of samples increases, so does the number of variables 
and the optimization procedure may become extremely costly. 
We shall discuss how to deal with the computational inefficiency later.

For the case when the number of target samples 
equals to the number of source samples; 
that is, $n_t = n_s$, 
the correspondence can be represented by 
a permutation matrix $\bP \in \{0,1\}^{n_s \times n_t}$. 
Element $[\bP]_{ij}=1$ 
if the source-domain sample $\bx^s_i$ 
corresponds to the target-domain sample $\bx^t_j$, and 0, otherwise. 
The permutation matrix $\bP$ has constraints 
$\sum_{i}[\bP]_{ij} = 1$ and $\sum_{j}[\bP]_{ij} = 1$  
for all $i \in \{1,2, \ldots, n_s\}$ and $j \in \{1,2, \ldots, n_t\}$. 
Hence, if $\bX^s \in \mathbb{R}^{n_s \times d}$ and 
$\bX^t \in \mathbb{R}^{n_t \times d}$ be the data matrix of the source-domain 
and the target-domain data, respectively, then $\bP\bX^t$ 
permutes the target-domain data matrix.

As soon as the correspondence is established, 
a linear or a non-linear mapping must be established 
between the target samples and the corresponding source samples. 
Non-linear mapping is involved when there is localized mapping for each sample, 
and it might also be required in case there is unequal domain shift of each class. 
The mapping operation should map the source-domain samples 
as close as possible to the corresponding target-domain samples. 
This process of finding a correspondence between these transformed source samples 
and target samples and then finding the mapping will continue iteratively till convergence.
This iterative method of alternately finding the correspondence and mapping 
is similar to feature registration in computer vision~\citep{chui2003new,besl1992method} 
but they have not been used or reformulated for unsupervised domain adaptation . 
In fact, the feature registration methods formulate 
the problem as a non-convex optimization. 
Consequently, these methods suffer from local minimum as in~\citep{besl1992method}, 
and the global optimization technique such as deterministic annealing~\citep{chui2003new} 
does not guarantee convergence. 
Thus, we propose to formulate it as a convex optimization problem to obtain correspondences as a global solution.
It is important to note that finding such global and unique solution 
to the correspondence accurately is more important because mapping 
with inaccurate correspondences will undoubtedly yield bad results. 

Formulating the proposed unsupervised domain adaptation problem
as a convex optimization problem
requires the correspondences to have the following properties: 
(a) {\em First-order similarity:} 
The corresponding target-domain samples should be as close as possible to 
the corresponding source-domain samples. 
This implies that we want to have the permuted target-domain data matrix 
$\bP\bX^t$ to be close to the source-domain data matrix $\bX^s$,
which translates to minimizing the Frobenius norm $||\bP\bX^t-\bX^s||^2_\mathcal{F}$
in the least-squares sense . 
(b) {\em Second-order similarity:} 
The corresponding target-domain neighborhood should be structurally 
similar to the corresponding source-domain neighborhood. 
This structural similarity can be expressed using graphs 
constructed from the source and target domains. 
Thus, if the two domains can be thought of as weighted undirected graphs $G^s, G^t$, 
structural similarity implies matching edges between the source and the target graphs. 
The edges of these graphs can be expressed using the adjacency matrices. 
If $\bD^s$ and $\bD^t$ are the adjacency matrices of $G^s$ and $G^t$, respectively, 
then these adjacency matrices can be found as,
\begin{align*}
[\bD^s]_{ij} = \text{exp}({-\frac{||\bx^s_i-\bx^s_j||^2_2}{{\sigma}_s^2}})
\end{align*}
\begin{align*}
[\bD^t]_{ij} = \text{exp}({-\frac{||\bx^t_i-\bx^t_j||^2_2}{{\sigma}_t^2}})
\end{align*} 
\begin{align*}
[\bD^s]_{ii}=[\bD^t]_{ii}=0,
\end{align*}
where ${\sigma}_s$ and ${\sigma}_t$ can be found heuristically 
as the mean sample-to-sample pairwise distance 
in the source and target domains, respectively.
For the second-order similarity, 
we want the permuted target domain adjacency matrix $\bP\bD^t\bP^{T}$ 
to be close to the source domain adjacency matrix (region) $\bD^s$,
where the superscript $T$ indicates a matrix transpose operation. 
We formulate it as equivalent to minimizing $||\bP\bD^t\bP^{T}-\bD^s||^2_\mathcal{F}$. 
While this cost term geometrically implies the cost of 
mis-matching edges in the constructed graphs, 
the first-order similarity term can be thought as the cost of mis-matching nodes. 
However, the second-order similarity cost term is bi-quadratic 
and we want to make it quadratic so that the cost-function is convex 
and we can apply convex optimization techniques to it. 
This can be done by post-multiplying $\bP\bD^t\bP^{T}-\bD^s$ by $\bP$. 
Using the permutation matrix properties  $\bP^{T}\bP=\bI$ (orthogonal) 
and $||\bA\bP||=||\bA||$ (norm-preserving), 
this transformation produces the cost function $||\bP\bD^t-\bD^s\bP||^2_\mathcal{F}$. 

Estimating the correspondence as a permutation matrix in this quadratic setting 
is NP-hard because of the combinatorial complexity of the constraint on $\bP$. 
We can relax the constraint on the correspondence matrix by converting it 
from a discrete to a continuous form. 
The norms (i.e., Frobenius) used in the cost/regularization terms 
will yield a convex minimization problem if we replace $\bP$ with a continuous constraint. 
Hence, if we relax the constraints on $\bP$ to allow for 
soft correspondences (i.e., replacing $\bP$ with $\bC$), 
then an element of $\bC$ matrix, $[\bC]_{ij}$, 
represents the probability that $\bx^s_i$ corresponds to $\bx^t_j$. 
This matrix $\bC$ is called doubly stochastic matrix 
{$\mathcal{D}_B=\{\bC \geq \bzero : \bC\mathbf{1}=\bC^{T}\mathbf{1}=\mathbf{1}\}$ . $\mathcal{D}_B$ represents a convex hull, containing all permutation matrices at its vertices}. (Birkhoff-von-Neumann theorem).

In addition to the graph-matching terms, 
we add a class-based regularization to the cost function
that exploits the labelled information of source-domain data. 
The group-lasso regularizer $\ell_2 - \ell_1$ norm term is equal to 
$\sum_j\sum_{c}||[\bC]_{\mathcal{I}_{c}j}||_2$, 
where $||\cdot||_2$ is the $\ell_2$ norm and $\mathcal{I}_{c}$ 
contains the indices of rows of $\bC$ 
corresponding to the source-domain samples of class $c$. In other words, $[\bC]_{\mathcal{I}_{c}j}$ is a vector consisting of elements $[\bC]_{ij}$, where $i^{th}$ source sample belongs to class $c$ and the $j^{th}$ sample is in the target domain.
Minimizing this group-lasso term ensures that a target-domain sample only 
corresponds to the source-domain samples that have the same label. 

It is important to note that the solution to the relaxed problem 
may not be equal or even close to the original discrete problem. 
Even then, the solution of the relaxed problem need not be projected onto 
the set of permutation matrices to get our final solution. 
This is because the graphs constructed using the source samples 
and the target samples are far from isomorphic for real datasets. 
Therefore, we do not expect exact matching between the nodes (samples) 
of each graph (domain) and soft correspondences may serve better. 
As an example, consider that a source sample $\bx^s_i$ 
is likely to correspond to both $\bx^t_j$ and $\bx^t_k$. 
In that case, it is more appropriate to have correspondences 
$[\bC]_{ij}=0.7$ and $[\bC]_{ik}=0.3$ assigned to the target samples, 
rather than the exact correspondences $[\bC]_{ij}=1$ and $[\bC]_{ik}=0$ or vice-versa. 
Thus, we can formulate our optimization problem of obtaining $\bC$ as follows:

\begin{align}
\label{prob0}
\min \limits_{\bC} f (\bC) = & ||\bC\bX^t-\bX^s||^2_\mathcal{F}/(n_sd)
+  \\
\lambda_s||\bC\bD^t-\bD^s\bC||^2_\mathcal{F} 
& + \lambda_g\sum_j\sum_{c}||[\bC]_{\mathcal{I}_{c}j}||_2 \nonumber \\
{\rm such~that~} ~ \bC \geq \bzero, & ~ \bC\mathbf{1}_{n_t}=\mathbf{1}_{n_s}, ~ {\rm and} ~\bC^T\mathbf{1}_{n_s}=\mathbf{1}_{n_t }, \nonumber 
\end{align}
where $\lambda_s$ and $\lambda_g$ are the parameters weighing 
the second-order similarity term and class-based regularization term, respectively; 
$\mathbf{1}_{n_s}$ and $\mathbf{1}_{n_t }$ are column vectors 
of size $n_s$ and $n_t$, respectively,
and the superscript $T$ indicates a matrix transpose operation.
The assumption that $n_t=n_s$ is strict 
and it needs to be relaxed to allow more realistic situations such as $n_t \neq n_s$.  
To analyze what modification is required to the optimization problem in Eq.~\eqref{prob0}, 
we explore further to understand the correspondences properly. 
In the case of $n_t=n_s$, 
we have one-to-one correspondences between each source sample and each target sample. 
However, for the case $n_t \neq n_s$, we must allow multiple correspondences. 
Initially, the constraint $\bC\mathbf{1}_{n_t}=\mathbf{1}_{n_s}$ implies that 
the sum of the correspondences of all the target samples to each source sample is one. 
The second equality constraint $\bC^{T}\mathbf{1}_{n_s}=\mathbf{1}_{n_t}$ implies 
that the sum of correspondences of all the source samples to each target sample is one. 
However, if $n_t \neq n_s$, the sum of correspondences of all the source samples 
to each target sample should increase proportionately by $\frac{n_s}{n_t}$ 
to allow for the multiple correspondences. 
This is reflected in the following optimization problem.\\[-0.1in]

\noindent
\emph{Problem UDA} 
\begin{align}
\label{prob1}
 \min \limits_{\bC} f (\bC)  & =  ||\bC\bX^t-\bX^s||^2_\mathcal{F}/(n_sd)  + 
 \lambda_s||\bC\bD^t - (\frac{n_t}{n_s})\bD^s\bC||^2_\mathcal{F} \\
 & \quad +  \lambda_g\sum_j\sum_{c}||[\bC]_{\mathcal{I}_{c}j}||_2   \nonumber \\
{\rm such~that} ~ \bC \geq \bzero,  ~ \bC\mathbf{1}_{n_t} &= \mathbf{1}_{n_s}, ~ {\rm and} ~\bC^T\mathbf{1}_{n_s}=(\frac{n_s}{n_t})\mathbf{1}_{n_t } \nonumber
\end{align}
for $n_t \neq n_s$.

\subsection{Correspondence-and-Mapping Problem Solution}
\textit{Problem UDA} is a constrained convex optimization problem 
and can easily be solved by interior-point methods~\citep{boyd2004convex}. 
In general, the time complexity of these interior-point-methods 
for conic programming is $O(N^{3.5})$, 
where $N$ is the total number of the variables~\citep{concomp}. 
If we have $n_s$ and $n_t$ as source and target samples, respectively, 
then the time complexity becomes $O(n_s^{3.5}n_t^{3.5})$. 
Also, the interior-point method is a second-order optimization method. 
Hence, it requires storage space of the Hessian, 
which is $O(N^{2})$ $\sim$ $O(n_s^{2}n_t^{2})$. 
This space complexity is more alarming and does not scale well 
with an increasing number of variables. 
If $n_t$ and $n_s$ are greater than 100 points, 
it results in memory/storage-deficiency problems in most personal computers. 
Thus, we need to employ a different optimization procedure 
so that the proposed UDA approach can be widely used without memory-deficiency problem. 
We could think of first-order methods of solving the constrained optimization problem, 
which require computing gradients but do not require storing the Hessians. 

First-order methods of solving the constrained optimization problem  
can be broadly classified into projected-gradient methods 
and conditional gradient (CG) methods~\citep{frank1956algorithm}.  
The projected-gradient method is similar to the normal gradient-descent method 
except that for each iteration, the iterate is projected back into the constraint set. 
Generally, the projected gradient-descent method enjoys 
the same convergence rate as the unconstrained gradient-descent method. 
However, for the projected gradient-descent method to be efficient, 
the projection step needs to be inexpensive. 
With an increasing number of variables, the projection step can become costly. 
Furthermore, the full gradient updating may destroy the structure of the solutions 
such as sparsity and low rank. 
The conditional gradient method, on the other hand, 
maintains the desirable structure of the solution 
such as sparsity by solving the successive linear minimization sub-problems 
over the convex constraint set. 
Since we expect our correspondence matrix $\bC$ to be sparse,
we shall employ the conditional gradient method for our problem.
In fact, \citet{jaggi2013revisiting} points out that convex optimization problems 
over convex hulls of atomic sets, which are relaxations of NP-hard problems 
are directly suitable for the conditional gradient method. 
This is similar to the way we formulate our problem by relaxing $\bP$ matrix to $\bC$. 
     
\begin{algorithm}[]
\SetAlgoLined
 \textbf{Given :} $\bC_0 \in \mathcal{D}$, $t=1$\\
 \textbf{Repeat}\\
 \quad $\bC_d = \text{arg}\min \limits_{\bC} \text{Tr}(\nabla_C f(\bC_0)^{T}\bC), \quad \text{such that} \quad \bC \in \mathcal{D}$ \\
 \quad $\bC_1 = \bC_0 + \alpha(\bC_d - \bC_0), \quad \text{for} \quad \alpha = \frac{2}{t+2}$ \\
 \quad $\bC_0=\bC_1$ \quad \text{and} \quad $t=t+1$ \\
 \textbf{Until} Convergence or Fixed Number of Iterations \\
 \textbf{Output :} $\bC_0 = \text{arg}\min \limits_{\bC} f(\bC) \quad \text{such that} \quad \bC \in \mathcal{D}$
\caption{Conditional Gradient Method (CG).}
\end{algorithm}

As described in the above \emph{Algorithm 1}
of the conditional gradient method, 
we have to solve the linear programming problem, $\min \limits_{\bC} \text{Tr}(\nabla_C f(\bC_0)^{T}\bC)$,
such that $\bC \in \mathcal{D}=\{\bC: \bC \geq \bzero, \bC\mathbf{1}_{n_t}
=\mathbf{1}_{n_s}, 
\bC^T\mathbf{1}_{n_s}=(\frac{n_s}{n_t})\mathbf{1}_{n_t }\}$. 
Here $\text{Tr}(\cdot)$ is the Trace operator. 
The gradient $\nabla_\bC f$ can be found from the equation:
\begin{equation}\
\label{grad}
\nabla_{\mathbf{C}} f  = \nabla_{\mathbf{C}}f_1 /(n_sd) +
\lambda_s\nabla_{\mathbf{C}}f_2  +
\lambda_g\nabla_{\mathbf{C}}f_3 ,
\end{equation}
where $f_1$, $f_2$, and $f_3$ are $||\bC\bX^t-\bX^s||^2_\mathcal{F}$, $||\bC\bD^t - (\frac{n_t}{n_s})\bD^s\bC||^2_\mathcal{F}$, 
and $\sum_j\sum_{c}||[\bC]_{\mathcal{I}_{c}j}||_2 $, respectively. 

{The gradients are obtained as follows. The derivation is given in the Appendix section}-
\begin{align*}
\nabla_{\mathbf{C}}f_1=
2(\bC\bX^{t}-\bX^{s})(\bX^t)^T
\end{align*}
\begin{align*}
\nabla_{\mathbf{C}}f_2=
2\bC\bD^{t}(\bD^t)^T-
2r\bD^{s}\bC(\bD^t)^T-2r(\bD^s)^T\bC\bD^{t} + 2r^{2}(\bD^s)^T\bD^{s}\bC
\end{align*}
where $r=\frac{n_t}{n_s}$ and
\begin{equation*}   
\frac{\partial f_3}{\partial [\bC]_{ij}}= 
\begin{cases}
 \frac{[\bC]_{ij}}{||[\bC]_{\mathcal{I}_{c}(i)j}||_2}~, & \text{if } ||[\bC]_{\mathcal{I}_{c}(i)j}||_2 \neq 0 ;\\
    0~,              & \text{otherwise;}
\end{cases}
\end{equation*}
Here, $c(i)$ is the class corresponding to the $i^{th}$ sample in the source domain and $\mathcal{I}_{c}(i)$ contains the indices of source samples belonging to class $c(i)$. After the gradient $\nabla_\bC f$ is found 
from $\nabla_\bC f_1$, $\nabla_\bC f_2$, $\nabla_\bC f_3$  using Eq.~\ref{grad}, 
we need to solve for the linear programming problem.

The linear programming problem can be solved easily using  
simplex methods used in solvers such as MOSEK~\citep{mosek2010mosek}. 
However, using such solvers would not make our method 
competitive in terms of time efficiency. 
Hence, we convert this linear programming problem 
into a \emph{min-cost flow} problem, 
which can then be solved very efficiently 
using a network simplex approach~\citep{kelly1991minimum}. 

Let the gradient $\nabla_\bC f(\bC_0)$ be $\bG/n_s$ 
and the correspondence matrix variable be $\bC=n_s\bT$. 
Then, the linear programming (LP) problem translates 
to $\min \limits_{\bT} \text{Tr}(\bG^{T}\bT)$
such that $\bT \geq \bzero, \bT\mathbf{1}_{n_t}=\mathbf{1}_{n_s}/n_s, 
\bT^T\mathbf{1}_{n_s}=\mathbf{1}_{n_t}/n_t$. 
This LP problem has an equivalence with the min-cost flow problem 
on the following graph:
\begin{itemize}
\item The graph is bipartite with $n_s$ source nodes and $n_t$ sink nodes.
\item The supply at each source node is $1/n_s$ and the demand at each sink node is $1/n_t$. 
\item Cost of the edge connecting the $i^{th}$ source node 
to the $j^{th}$ sink node is given by $[\bG]_{ij}$. 
Capacity of each edge is $\infty$.
\end{itemize}
Using this configuration, 
the min-cost flow problem is solved using the network simplex. 
Details of the network-simplex method is omitted 
and one can refer~\citep{kelly1991minimum}. 
The network simplex method is an implementation of 
the traditional simplex method for LP problems, 
where all the intermediate operations are performed on graphs. 
Due to the structure of min-cost flow problems, 
network-simplex methods provide results significantly 
faster than traditional simplex methods. 
Using this network-simplex method, 
we obtain the solution $\bT^{*}$, 
where $[\bT^{*}]_{ij}$ is the flow obtained on the edge connecting the $i^{th}$ source node 
to the $j^{th}$ sink node. 
From that, we obtain $\bC_d=n_s\bT^{*}$ and proceed with that iteration 
of conditional gradient (CG) method as in \emph{Algorithm 1}.
In the above CG method, we also need an initial $\bC_0$ and 
$\bC_0$ can be defined as the solution to the LP problem, 
$\min \limits_{\bC}\text{Tr}(\bD^{T}\bC)$ 
such that $\bC \in \mathcal{D}=\{\bC : \bC \geq 0, \bC\mathbf{1}_{n_t}=\mathbf{1}_{n_s}, 
\bC^T\mathbf{1}_{n_s}=(\frac{n_s}{n_t})\mathbf{1}_{n_t }\}$, 
where $[\bD]_{ij}=||\bx^s_i-\bx^t_j||_2$. 
This is also solved by the network simplex approach 
after converting this LP problem into its equivalent min-cost flow problem 
as described previously. 
After we obtain $\bC^{*}$ from the CG algorithm, 
it is then used to find the corresponding target samples $\bX^t_{c}=\bC^{*}\bX^t$. 
Then, the mapping $\bM(\cdot)$ from the source domain to the target domain 
is found by solving the following regression problem 
$\bM(\cdot) : \mathcal{X}^s \rightarrow \mathcal{X}^t$, 
with each row of $\bX^s$ as an input data sample and 
the corresponding row of $\bX^t_c$ as an output data sample. 
The choice of regressors can be linear functions, neural networks, 
and kernel machines with proper regularization. 
Once the mapping $\bM^*(\cdot)$ is found out, 
a source-domain sample $\bx^s$ can be mapped to the target domain 
by applying $\bM^*(\bx^s)$. 
This completes one iteration of finding the correspondence and the mapping. 
For the next cycle, we solve Problem UDA with the mapped source samples 
as $\bX^s$ and subsequently find the new mapping. 
The number of iterations $N_T$ of alternatively finding correspondence 
and mapping is an user-defined variable. 
The full domain adaptation algorithm is outlined in \emph{Algorithm 2}

\begin{algorithm}[H]
\SetAlgoLined
 \textbf{Given :} Source Labelled Data $\bX^s$ and $\bY^s$, and Target Unlabelled Data $\bX^t$\\
 \textbf{Parameters :} $\lambda_s, \lambda_g, N_T$\\
 \textbf{Initialize :} $t=0$\\
 \textbf{Repeat}\\
 \quad $\bC^* = \text{argmin} f (\bC) \quad \text{such that} \quad \bC \in \mathcal{D}$ \quad  (Find Correspondence using \hspace*{3.2in} CG method)\\
 \quad Regress $\bM(\cdot) \quad \text{s.t.} \quad \bX^s \xrightarrow[]{\textbf{M}}\bC^*\bX^t$ \quad (Find Mapping) \\
 \quad Map $\bX^s=\bM(\bX^s)$ \quad and \quad $t=t+1$\\
 \textbf{Until} $t= N_T$ \\
 \textbf{Output :} Adapted Source Data $\bX^s$,$\bY^s$ to learn classifier.
\caption{Unsupervised Domain Adaptation using dataset registration.}
\end{algorithm}

\section{Experimental Results and Discussions}
To evaluate and validate the proposed sample-sample correspondence 
and mapping method for unsupervised domain adaptation,
computer simulations were performed on a toy dataset 
and then on image classification and sentiment classification tasks. 
Our results were compared with previous published methods. 
For comparisons, we used the reported accuracies or conduct experiments 
with the available source code. 
Since we are dealing with unsupervised domain adaptation, 
it is not possible to cross-validate 
our hyper-parameters $\lambda_s$,   $\lambda_g$, and $N_T$. 
Unless explicitly mentioned, we reported the best results obtained 
over the hyper-parameter ranges $\lambda_s$ and $\lambda_g$ in  $\{10^{-3},10^{-2},10^{-1},10^{0},10^{1},10^{2},10^{3}\}$ and $N_T=1$. 
In our simulations, we found that using $N_T > 1$ 
only provides a tiny bump in performance or no improvement in performance at all. 
This is because the source samples have already been 
transformed close to the target samples 
and further transformation does not affect recognition accuracies. 
After the correspondence was found, 
we considered mapping between the corresponding samples. 
For the mapping, we used a linear mapping $\mathbf{W} \in \mathbb{R}^{d \times d}$ 
with a regularization of 0.001. $d$ is the dimension of the feature space in which the data lies.
\begin{figure*}[bht]
\centering
\includegraphics[width=5.5in]{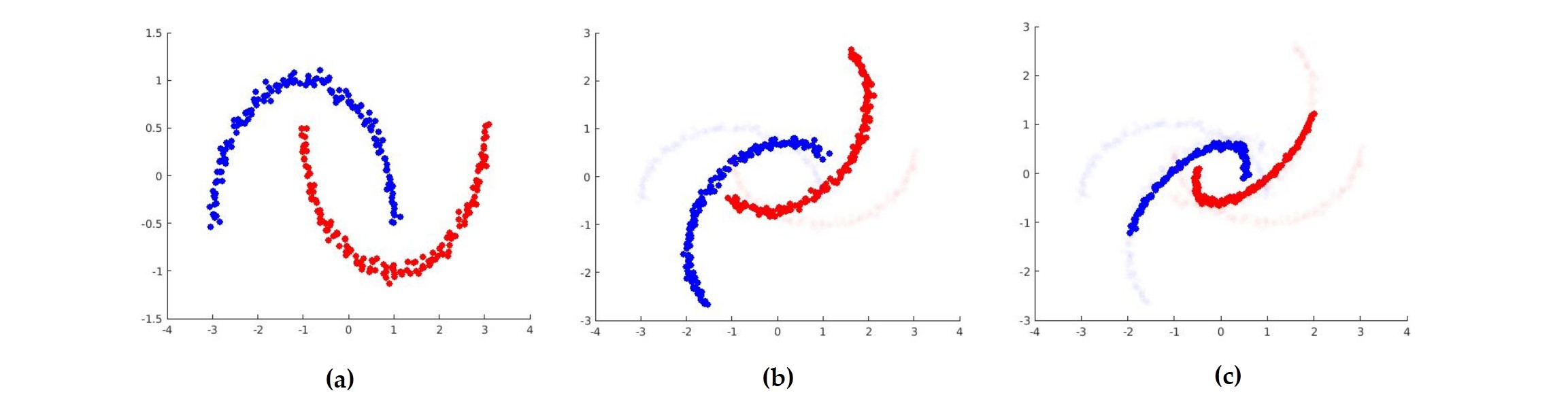}
\caption{(a) Source-domain data. 
(b) Target-domain data consists of a 50-degree rotation of the source-domain data. 
(c) Transformed source-domain data is now aligned with the target-domain data.}
\label{toydata}
\end{figure*}
\subsection{Toy Dataset: Two interleaving moons}
For the first experiment, 
we used the synthetic dataset of interleaving moons 
previously used in~\citep{courty2016optimal,germain2013pac}. 
The dataset consists of 2 domains. 
The source domain consists of 2 entangled moon's data. 
Each moon is associated with each class. 
The target domain consists of applying a rotation to the source domain. 
This can be considered as a domain-adaptation problem 
with increasing rotation angle implying increasing difficulty 
of the domain-adaptation problem. 
Since the problem is low dimensional, 
it allowed us to visualize the effect of our domain-adaptation method appropriately. 
Figures~\ref{toydata}(a) and~\ref{toydata}(b) 
show an example of the source-domain data and  the target-domain data respectively, 
and Fig.~\ref{toydata}(c) shows the adapted source-domain data 
using the proposed approach. 
The results showed that the transformed source domain becomes close to the target domain.

For testing on this toy dataset, 
we used the same experimental protocol as in~\citep{courty2016optimal,germain2013pac}. 
We sampled 150 instances from the source domain and 
the same number of examples from the target domain. 
The test data was obtained by sampling 1000 examples 
from the target domain distribution. 
The classifier used is an SVM with a Gaussian kernel, 
whose parameters are set by 5-fold cross-validation.
The experiments were conducted over 10 trials and the mean accuracy was reported. 
{At this juncture, it is important to note that choosing the classifier for domain adaptation is important. For example, the two classes in the interleaving moon dataset are not linearly separable at all. So, a linear kernel SVM would not classify the moons accurately and it would result in poor performance in the target domain as well. That is why we need a Gaussian Kernel SVM. So, we have to make sure that we choose a classifier that works well with the source dataset in the first place.}
\begin{table}[]
\centering
\caption{Accuracy results over 10 trials for the toy dataset domain-adaptation problem for varying degree of rotation between source and target domain}\label{tab:toytable}
\begin{tabular}{c c c c c c c c}
\hline
\textbf{Angle ($^{\circ}$)} & 10 & 20    & 30    & 40    & 50    & 70    & 90    \\ \hline
SVM-NA                  & 100  & 89.6 & 76.0 & 68.8 & 60.0   & 23.6 & 17.2\\ 
DASVM                  & 100  & 100 & 74.1  & 71.6 & 66.6 & 25.3 & 18.0  \\ 
PBDA                   & 100  & 90.6 & 89.7 & 77.5 & 59.8 & 37.4 & 31.3 \\ 
OT-exact                & 100  & 97.2 & 93.5 & 89.1 & 79.4 & 61.6 & 49.3 \\ 
OT-IT                   & 100  & 99.3 & 94.6 & 89.8 & 87.9 & 60.2 & 49.2 \\ 
OT-GL                   & 100  & 100     & \textbf{100}     & \textbf{98.7} & 81.4 & 62.2 & 49.2 \\ 
OT-Laplace              & 100  & 100     & 99.6 & 93.8 & 79.9  & 59.8 & 47.6 \\ 
Ours                    & 100  & 100     & 96    & 87.4    & \textbf{83.9} & \textbf{78.4} & \textbf{72.2} \\ \hline
\end{tabular}

\end{table}

{We} compared our results with the DA-SVM~\citep{bruzzone2010domain}- a domain-adaptive support vector machine approach, 
PBDA~\citep{germain2013pac}- which is a PAC-Bayesian based domain adaptation method, and different versions of the optimal transport approach~\citep{courty2016optimal}. OT-exact is the basic optimal transport approach. OT-IT is the information theoretic version with entropy regularization. OT-GL and OT-Laplace has additional group and graph based regularization, respectively.
From our results in Table~\ref{tab:toytable}, 
we see that for low rotation angles, the OT-GL-based method dominates 
and our proposed method yields satisfactory results. 
But for higher angles ($\geq$ 50 $^{\circ}$), 
our proposed method clearly dominates by a large margin.  
This is because we have taken into consideration second-order 
structural similarity information. 
For higher-rotation angles, 
the point-to-point sample distance is high. 
However, similar structures in the source and target domains 
can still correspond to each other. 
In other words, the adjacency matrices, 
which depend on relative distances between samples, 
can still be matched and do not depend on higher rotation angles 
between the source and target domains. 
That is why our proposed method out-performed other methods 
for large discrepancies between the source and target distributions.

We further provided the time comparison between the network simplex method (N-S) 
and MOSEK (M) for increasing number of samples 
of the toy dataset in Table~\ref{toytime}. 
Results showed that the network simplex method is very fast compared to 
a general purpose linear programming solver like MOSEK.

\begin{table}[]
\centering
\caption{Time comparison (in seconds) of the two solvers for increasing sample size. 
The sample size is the number of samples per class per domain 
of the interleaving moon toy dataset. 
The target domain has a rotation of $50^{\circ}$ with the source domain. 
We use $N_T=1$. 
Implementation was in MATLAB in a workstation with {Intel}® Xeon(R) CPU E5-2630 v2 and 40 GB RAM. Results are reported over 10 trials.}\label{toytime}
\begin{tabular}{c c c c c c c c c}
\hline
$n$ & 25    & 50    & 75     & 100    & 125    & 150    & 175    & 200    \\ \hline
M               & 62.1 & 83.1 & 103.4 & 128.5 & 387.7 & 680.1 & 1028.3 & 1577.6 \\ 
N-S     & 1.5  & 4  & 6.9   & 10.1  & 16.9  & 23.5  & 31.2  & 41.3  \\ \hline
\end{tabular}
\end{table}

\begin{figure*}[htb]
\centering
\includegraphics[width=5.5in]{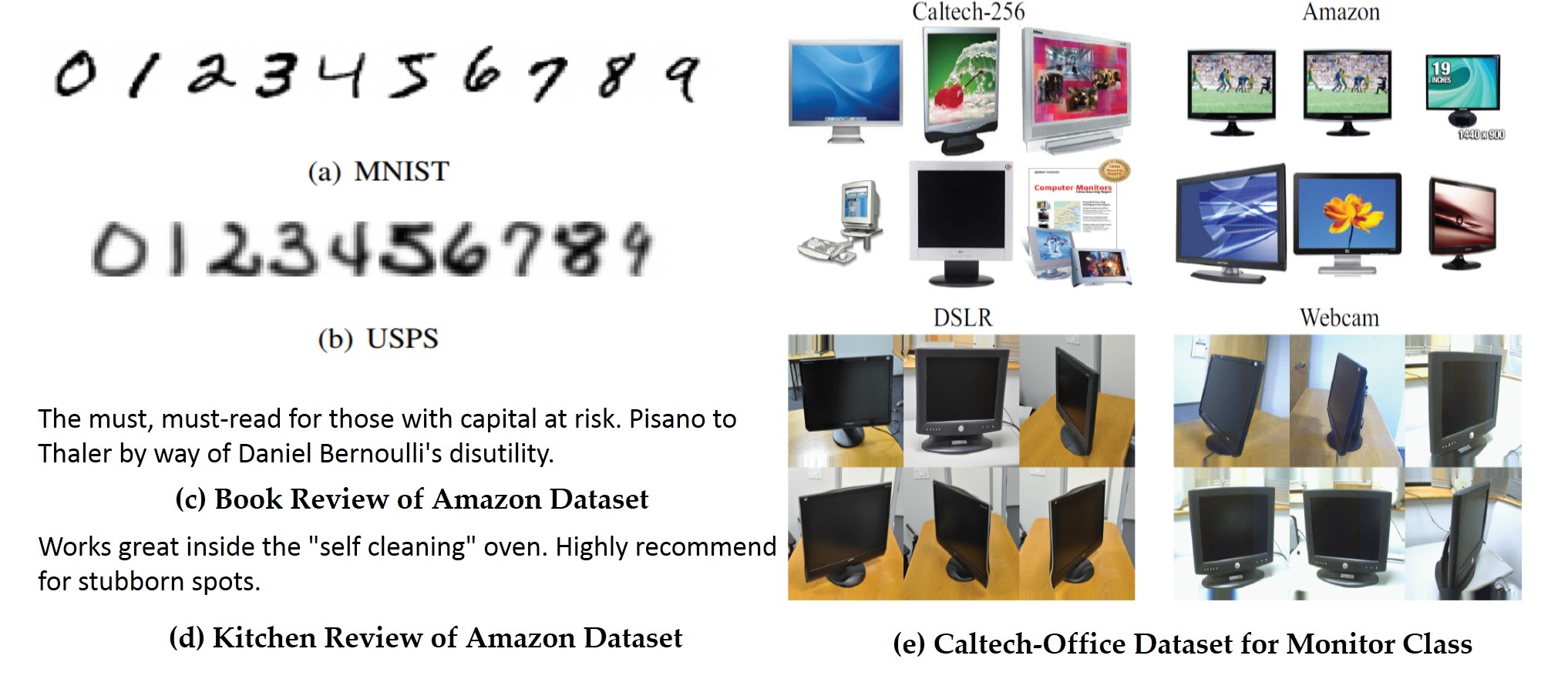}
\caption{Instances of the real dataset used. 
At the top left, we see that USPS has the worse resolution 
compared to MNIST handwriting dataset. 
At the bottom left, we have instances of the Amazon review dataset. 
There is a shift in textual domain when reviewing for different products. 
On the right, we have the Caltech-Office dataset and we see that there are differences 
in illumination, quality, pose, presence/absence of background across different domains.}
\label{dataset}
\end{figure*}
\subsection{Real Dataset: Image Classification}
We next evaluated the proposed method on image classification tasks. 
The image classification tasks that we considered 
were digit recognition and object recognition. 
The classifier used was 1-NN (Nearest Neighbor). {1-NN is used for experiments with images because it does not require cross-validating hyper-parameters and has been used in previous work as well} \citep{courty2016optimal,gong2012geodesic}
The 1-NN classifier is trained on the transformed source-domain data 
and tested on the target-domain data. Instances of the image dataset are shown in Fig. \ref{dataset} (a),(b) and (e). {Generally, we cannot directly cross-validate our hyper-parameters $\lambda_s$ and $\lambda_g$ on the unlabelled target domain data making it impractical for real-world applications. However, for practical transfer learning purposes, a reverse validation (RV)
 technique}\citep{zhong2010cross} {was developed for tuning the hyper-parameters. We have carried out experiments with a variant of the method to tune $\lambda_s$ and $\lambda_g$ for our UDA approach.}. 

{For a particular hyper-parameter configuration, we divide the source domain data into $K$ folds. We use one of the folds as the validation set. The remaining source data and the whole target data are used for domain adaptation. The classifier trained using the adapted source data is used to generate pseudo-labels for the target data. Another classifier is trained using the target domain data and its pseudo-labels. This classifier is then tested on the held-out source domain data after adaptation. The accuracy obtained is repeated and averaged over all the $K$ folds. This reverse-validation approach is repeated over all hyper-parameter configurations. The optimal hyper-parameter configuration is the one with the best average validation accuracy. Using the obtained optimal hyper-parameter configuration, we then carry out domain adaptation over all the source and target domain data and report the accuracy over the target domain dataset. We used $K=5$ folds for all the real-data experiments. Thus, we showed the results using this RV approach in addition to the best obtained results over the hyper-parameters. In majority of the cases in }Tables \ref{surf},\ref{decaf6},\ref{decaf7} and \ref{sentitab} {we would see that the result obtained using the reverse validation approach matches the best obtained results suggesting that the hyper-parameters can be automatically tuned successfully.}
\subsubsection{Digit Recognition} 
For the source and target domains, 
we used 2 datasets -- USPS (U) and MNIST (M). 
These datasets have 10 classes in common (0-9). 
The dataset consists of randomly sampling 1800 and 2000 images 
from USPS and MNIST, respectively. 
The MNIST digits have $28 \times 28$ resolution and the USPS $16 \times 16$. 
The MNIST images were then resized to the same resolution as that of USPS. 
The grey levels were then normalized to obtain a common 256-dimensional 
feature space for both domains.
\subsubsection{Object Recognition}
For object recognition, 
we used the popular Caltech-Office 
dataset~\citep{gong2012geodesic,gopalan2011domain,saenko2010adapting,
zheng2012grassmann,courty2016optimal}. 
This domain-adaptation dataset consists of images from 4 different domains: 
\emph{Amazon} (A) (E-commerce), \emph{Caltech-256}~\citep{griffin2007caltech} (C) 
(a repository of images), \emph{Webcam} (W) (webcam images), 
and \emph{DSLR} (D) (images taken using DSLR camera). 
The differences between domains are due to the differences in quality, illumination, pose 
and also the presence and absence of backgrounds. 
The features used are the shallow SURF features~\citep{surf} 
and deep-learning feature sets~\citep{decaf} -- \emph{decaf6} and \emph{decaf7}. 
The SURF descriptors represent each image as a 800-bin histogram. 
The histogram is first normalized to represent a probability 
and then reduced to standard z-scores. 
On the other hand, the deep-learning feature sets, \emph{decaf6} and \emph{decaf7}, 
are extracted as the sparse activation of the neurons from 
the fully connected 6th and 7th layers of convolutional network 
trained on imageNet and fine tuned on our task. 
The features are 4096-dimensional.

For our experiments, 
we considered a random selection of 20 samples per class 
(with the exception of 8 samples per class for the DSLR domain) for the source domain. 
The target-domain data is split equally. 
One half of the target-domain data is used for domain adaptation 
and the other half is used for testing. 
This is in accordance with the protocol followed in~\citep{courty2016optimal}. 
The accuracy is reported on the test data over 10 trials of the experiment.

We compared our approach against (a) the no adaptation baseline (NA), 
which consists of using the original classifier without adaptation; 
(b) Geodesic Flow Kernel (GFK)~\citep{gong2012geodesic}; 
(c) Transfer Subspace Learning (TSL)~\citep{si2010bregman}, 
which minimizes the Bregman divergence between low-dimensional embeddings 
of the source and target domains; 
(d) Joint Distribution Adaptation (JDA)~\citep{long2013transfer}, 
which jointly adapts both marginal and conditional distributions 
along with dimensionality reduction; 
(e) Optimal Transport~\citep{courty2016optimal} with the information-theoretic (OT-IT) 
and group-lasso version (OT-GL).  
Among all these methods, TSL and JDA are moment-matching methods 
while OT-IT, OT-GL and ours are sample-matching methods. 

The best performing method for each domain-adaptation problem is highlighted in bold.
From Table~\ref{surf}, 
we see that in almost all the cases, 
the OT-GL and our proposed method dominated over other methods, 
suggesting that sample-matching methods perform better than moment-matching methods. 
For the handwritten digit recognition tasks (U $\rightarrow$ M and M $\rightarrow$ U), 
our proposed method clearly out-performs GFK, TSL and JDA, 
but is slightly out-performed by OT-GL. 
This might be because the handwritten digit datasets U and M 
do not contain enough structurally similar regions 
to exploit the second-order similarity cost term. 
For the Office-Caltech dataset, 
the only time our proposed method was beaten by 
a moment-matching method was W $\rightarrow$ D, though by a slight amount. 
This is because W and D are closest pair of domains and using sample-based matching 
does not have outright advantage over moment-matching. 
The fact that W and D have the closest pair of domains is evident 
form the NA accuracy of 53.62, 
which is the best among NA accuracies of the Office-Caltech domain-adaptation tasks. 

{We have performed a runtime comparison in terms of the CPU time in seconds of our method with other methods and have shown the results in} Table \ref{time}. {The experiments performed are over the same dataset as used in Table} \ref{surf}. {From Table} \ref{time}, {we see that local methods like OT-GL and our method generally take more time than moment-matching method like JDA. Our method takes more time compared with OT-GL because of time taken in constructing adjacency matrices for the second order cost term. Overall, the  time taken for domain adaptation between USPS and MNIST datasets is more because they contain relatively larger number of samples, compared to the Office-Caltech dataset.}
\begin{table}[]
\centering
\caption{{Domain-adaptation results for digit recognition using USPS and MNIST datasets and object recognition with the Office-Caltech dataset using SURF features.}}
\label{surf}
\begin{tabular}{c c c c c c c c}
\hline
\textbf{Tasks}    & NA    & GFK   & TSL   & JDA   & OT-GL & Ours & Ours (RV)\\ \hline
U $\rightarrow$ M & 39.00 & 44.16 & 40.66 & 54.52 & \textbf{57.85} & 56.90 & 56.90   \\ 
M $\rightarrow$ U & 58.33 & 60.96 & 53.79 & 60.09 & \textbf{69.96} & 68.44 & 66.24  \\ 
C $\rightarrow$ A & 20.54 & 35.29 & 45.25 & 40.73 & 44.17 & \textbf{46.67}  & \textbf{46.67} \\ 
C $\rightarrow$ W & 18.94 & 31.72 & 37.35 & 33.44 & 38.94 & \textbf{39.48} & \textbf{39.48}   \\ 
C $\rightarrow$ D & 19.62 & 35.62 & 39.25 & 39.75 & \textbf{44.50} & 42.88  & 40.12 \\ 
A $\rightarrow$ C & 22.25 & 32.87 & 38.46 & 33.99 & 34.57 & \textbf{38.51}  & \textbf{38.51}  \\ 
A $\rightarrow$ W & 23.51 & 32.05 & 35.70 & 36.03 & 37.02 & \textbf{38.69}  & \textbf{38.69}  \\ 
A $\rightarrow$ D & 20.38 & 30.12 & 32.62 & 32.62 & \textbf{38.88} & 36.12 & 36.12   \\ 
W $\rightarrow$ C & 19.29 & 27.75 & 29.02 & 31.81 & \textbf{35.98} & 33.81 & 32.83   \\ 
W $\rightarrow$ A & 23.19 & 33.35 & 34.94 & 31.48 & \textbf{39.35} & 37.69 & 37.69 \\ 
W $\rightarrow$ D & 53.62 & 79.25 & 80.50 & \textbf{84.25} & 84.00 & 84.10 & 84.10 \\ 
D $\rightarrow$ C & 23.97 & 29.50 & 31.03 & 29.84 & 32.38 & \textbf{32.78}  & \textbf{32.78} \\ 
D $\rightarrow$ A & 27.10 & 32.98 & 36.67 & 32.85 & 37.17 & \textbf{38.33} & 37.61 \\ 
D $\rightarrow$ W & 51.26 & 69.67 & 77.48 & 80.00 & 81.06 & \textbf{81.12} & \textbf{81.12}   \\ \hline
\end{tabular}

\end{table}

\begin{table}[]
\centering
\caption{{CPU time (seconds) comparison of different domain adaptation algorithms.}}
\label{time}
\begin{tabular}{c c c c c c c}
\hline
\textbf{Task}  & NA   & GFK  & TSL    & JDA   & OT-GL  & Ours   \\ \hline
U$\rightarrow$M & 1.24 & 2.62 & 567.8  & 82.34 & 171.84 & 201.23 \\
M$\rightarrow$U & 1.13 & 2.43 & 522.37 & 81.13 & 168.23 & 196.15 \\
C$\rightarrow$A & 0.46 & 2.6  & 382.98 & 41.6  & 85.95  & 99.9   \\
C$\rightarrow$W & 0.24 & 1.45 & 157.52 & 37.89 & 78.73  & 101.1  \\
C$\rightarrow$D & 0.36 & 1.35 & 117.81 & 37.33 & 61.17  & 63.38  \\
A$\rightarrow$C & 0.54 & 2.69 & 462.12 & 40.11 & 105.87 & 126.18 \\
A$\rightarrow$W & 0.39 & 1.47 & 153.95 & 37.63 & 86.12  & 100.21 \\
A$\rightarrow$D & 0.42 & 1.31 & 115.87 & 36.82 & 69.29  & 82.1   \\
W$\rightarrow$C & 0.33 & 2.92 & 461.1  & 42.39 & 98.26  & 111.2  \\
W$\rightarrow$A & 0.61 & 2.52 & 388.23 & 41.64 & 94.38  & 101.45 \\
W$\rightarrow$D & 0.34 & 1.37 & 117.47 & 37.9  & 76.5   & 79.25  \\
D$\rightarrow$C             & 0.45 & 2.36 & 364.13 & 39.75 & 106.21 & 118.12 \\
D$\rightarrow$A             & 0.43 & 2.14 & 310.18 & 41.24 & 98.41  & 115.35 \\
D$\rightarrow$W             & 0.24 & 1.05 & 93.73  & 34.62 & 76.23  & 88.69 \\ \hline
\end{tabular}
\end{table}
\begin{table}[]
\centering
\caption{{Domain-adaptation results for the Office-Caltech dataset using \emph{decaf6} features.}}
\label{decaf6}
\begin{tabular}{c c c c c c c}
\hline
\textbf{Task} & NA    & JDA   & OT\_IT & OT-GL          & Ours & Ours(RV)          \\ \hline
C$\rightarrow$A            & 79.25 & 88.04 & 88.69  & \textbf{92.08} & 91.92 & 89.91         \\ 
C$\rightarrow$W            & 48.61 & 79.60 & 75.17  & \textbf{84.17}          &  83.58  & 81.23            \\ 
C$\rightarrow$D       & 62.75 & 84.12 & 83.38  & 87.25          &  \textbf{87.50}  & \textbf{87.50}            \\ 
A$\rightarrow$C            & 64.66 & 81.28 & 81.65  & 85.51          & \textbf{86.67} & 85.63               \\ 
A$\rightarrow$W            & 51.39 & 80.33 & 78.94  & \textbf{83.05} & 81.39 & 81.39         \\ 
A$\rightarrow$D            & 60.38 & 86.25 & 85.88  & 85.00          & \textbf{87.12} & \textbf{87.12}       \\ 
W$\rightarrow$C            & 58.17 & 81.97 & 74.80  & 81.45          & \textbf{82.13} & 81.64      \\ 
W$\rightarrow$A            & 61.15 & 90.19 & 80.96  & \textbf{90.62}          & 88.87 & 88.87       \\ 
W$\rightarrow$D            & 97.50 & 98.88 & 95.62  & 96.25          & \textbf{98.95} & \textbf{98.95}\\ 
D$\rightarrow$C            & 52.13 & 81.13 & 77.71  & \textbf{84.11}          & 83.72  & 83.72        \\ 
D$\rightarrow$A            & 60.71 & 91.31 & 87.15  & 92.31          & \textbf{92.65}  & \textbf{92.65}          \\ 
D$\rightarrow$W            & 85.70 & \textbf{97.48} & 93.77  & 96.29          & 96.69    & 96.13      \\ \hline
\end{tabular}
\end{table}

\begin{table}[]
\centering
\caption{{Domain-adaptation results for the Office-Caltech dataset using \emph{decaf7} features.}}
\label{decaf7}
\begin{tabular}{c c c c c c c}
\hline
\textbf{Task} & NA    & JDA   & OT-IT & OT-GL          & Ours & Ours(RV)         \\ \hline
C$\rightarrow$A            & 85.27 & 89.63 & 91.56 & \textbf{92.15} & 91.85 & 91.85          \\
C$\rightarrow$W            & 65.23 & 79.80 & 82.19 & 83.84          & \textbf{85.36} & \textbf{85.36}              \\ 
C$\rightarrow$D            & 75.38 & 85.00 & 85.00 & 85.38          & \textbf{85.88} & \textbf{85.88}              \\ 
A$\rightarrow$C            & 72.80 & 82.59 & 84.22 & \textbf{87.16}          & 86.67  & 85.39         \\ 
A$\rightarrow$W            & 63.64 & 83.05 & 81.52 & 84.50 & \textbf{86.09} & 85.36         \\ 
A$\rightarrow$D            & 75.25 & 85.50 & 86.62 & 85.25          & \textbf{87.37} & \textbf{87.37}              \\ 
W$\rightarrow$C            & 69.17 & 79.84 & 81.74 & \textbf{83.71}  & 82.80 & 82.80             \\ 
W$\rightarrow$A            & 72.96 & 90.94 & 88.31 & \textbf{91.98}   & 90.15 & 89.31            \\ 
W$\rightarrow$D            & 98.50 & 98.88 & 98.38 & 91.38          & \textbf{99.00} & \textbf{99.00} \\ 
D$\rightarrow$C            & 65.23 & 81.21 & 82.02 & \textbf{84.93}          & 82.20  & 82.20           \\ 
D$\rightarrow$A            & 75.46 & 91.92 & 92.15 & \textbf{92.92}          & 92.60  & 92.15           \\ 
D$\rightarrow$W            & 92.25 & 97.02 & 96.62 & 94.17          &  \textbf{97.10} & \textbf{97.10}    \\ \hline
\end{tabular}
\end{table}

\begin{figure}[]
\centering
\includegraphics[width=12cm]{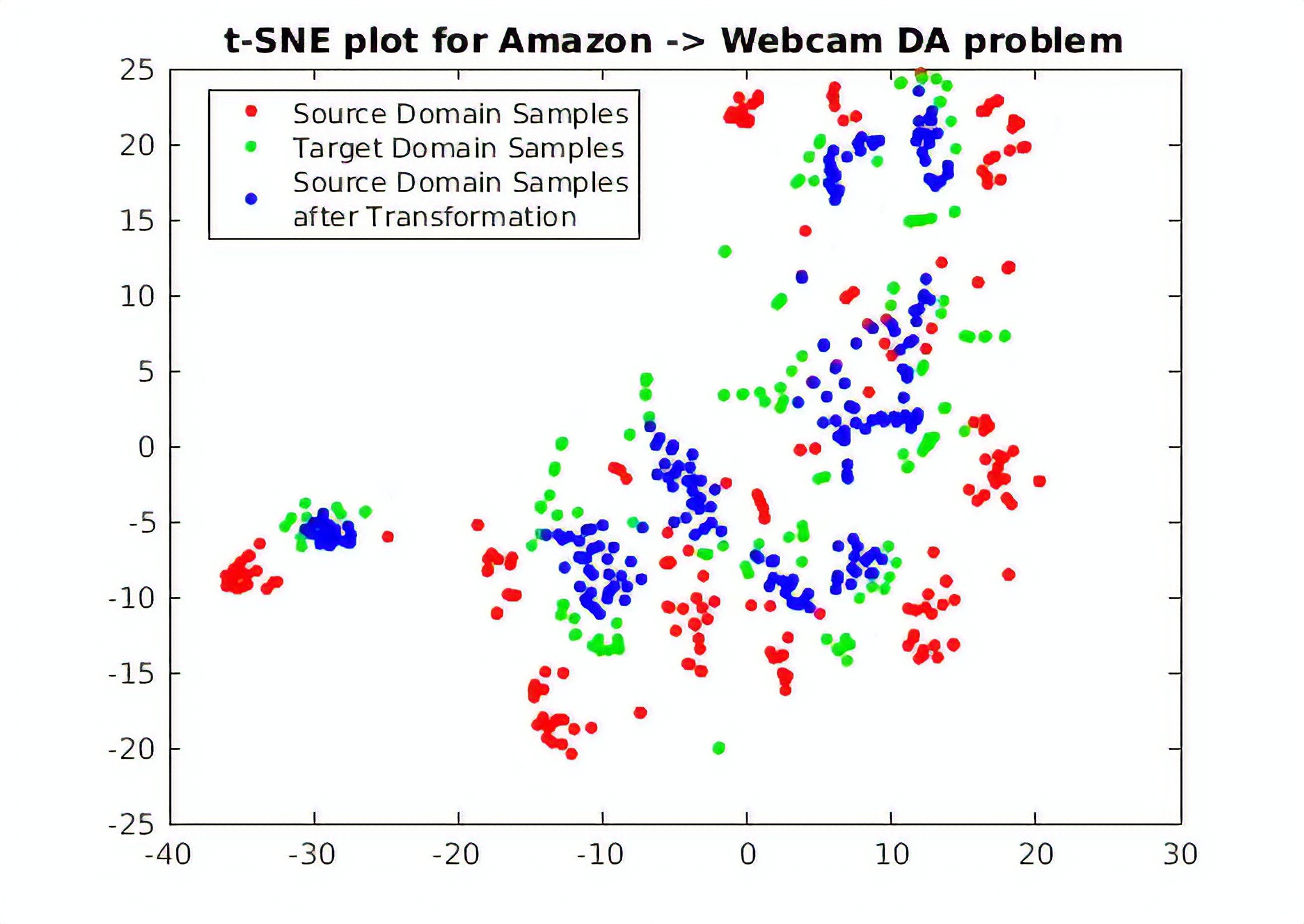}
\caption{t-SNE~\citep{maaten2008visualizing} visualization of  a single trial of Amazon to Webcam DA problem using \emph{decaf6} features.}
\label{tsne}
\end{figure}

We have also reported the results of Office-Caltech dataset 
using \emph{decaf6} and \emph{decaf7} features 
in Tables~\ref{decaf6} and~\ref{decaf7}, respectively. 
The baseline performance of the deep-learning features 
are better than SURF features 
because they are more robust and contain higher-level representations. 
Expectedly, the \emph{decaf7} features have better baseline performance 
than \emph{decaf6} features. 
However, DA methods can further increase performance over the robust deep features. 
In Tables~\ref{decaf6} and~\ref{decaf7}, 
we see that our proposed method dominates over JDA and OT-IT 
but is in close competition with OT-GL. 
We also noted that using \emph{decaf7} instead of \emph{decaf6}
creates only a small incremental improvement in performance 
because most of the adaptation has already been performed 
by our proposed domain-adaptation method.
As seen in Fig.~\ref{tsne}, the source-domain samples are transformed 
to be near the target-domain samples using our proposed method.  
Therefore, we expect a classifier trained on the transformed source samples 
to perform better on the target-domain data.

\begin{figure}[]
\centering
\includegraphics[width=8cm]{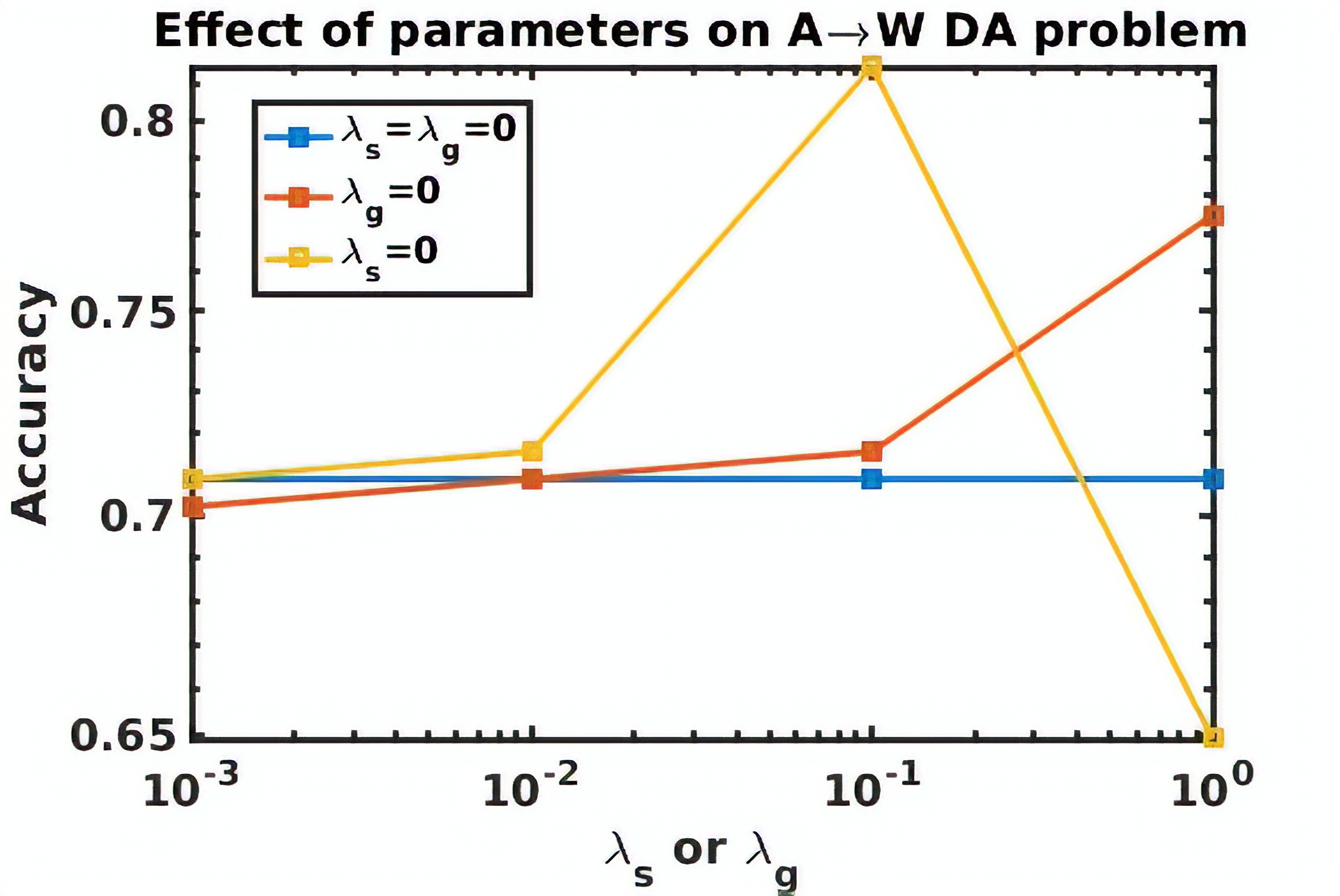}
\caption{Effect of varying regularization parameters $\lambda_s$ and $\lambda_g$ 
on the accuracy of Amazon (source domain) to Webcam (target domain) 
visual domain-adaptation problem for fixed $N_T=1$.}
\label{accu}
\end{figure}

\begin{figure}[]
\centering
\includegraphics[width=5.5in]{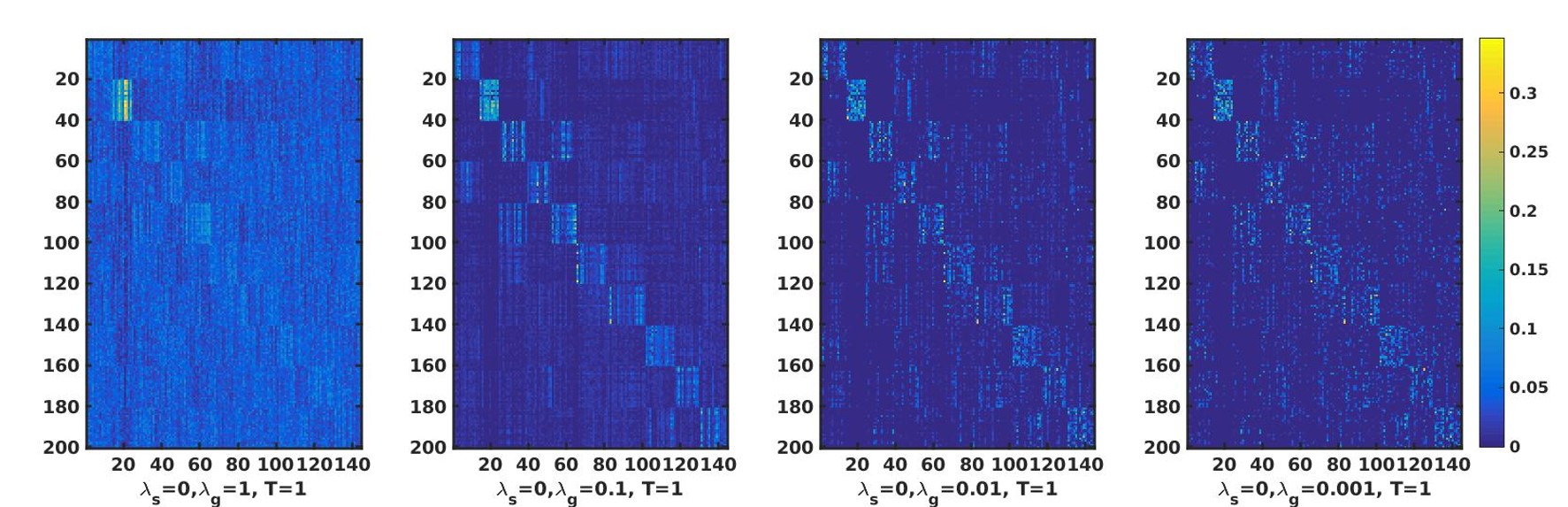}
\caption{The optimal correspondence matrix $\bC$ for 4 different parameter settings visualized as a colormap, with $\lambda_s=0, N_T=1$. 
The task involved was the Amazon to Webcam domain adaptation.}
\label{HM}
\end{figure}

We have also studied the effects of varying the regularization parameters 
on domain-adaptation performance. 
In {Fig.}~\ref{accu}, 
the blue line shows the accuracy when both $\lambda_s=\lambda_g=0$. 
When $\lambda_s=0$, best performance is obtained for $\lambda_g=0.1$. 
When $\lambda_g=0$, best performance is obtained for $\lambda_g=1$. 
For $\lambda_s,\lambda_g  > 1$, 
performance degrades (not shown) because we have put excess weight on 
the regularization terms of second-order structural similarity 
and group-lasso than on the first-order point-wise similarity cost term. {Thus, the presence of second-order and regularization term, weighted in the right amount is justified as it improves performance over when only the first-order term is present.}
We have also studied the effect of group-lasso regularization parameter ($\lambda_g$) 
on the quality of the correspondence matrix $\mathbf{C}$ 
obtained for a domain-adaptation task. 
Visually the second plot from the left in {Fig.}~\ref{HM} 
appears to discriminate the 10 classes best. 
Accordingly, this parameter configuration ($\lambda_s=0,
\lambda_g=0.1, N_T=1$) realizes the best performance 
as shown in the previous {Fig.}~\ref{accu}.

\subsection{Real Dataset: Sentiment Classification}
We have also evaluated our proposed method on sentiment classification 
using the standard Amazon review dataset~\citep{blitzer2007biographies}. 
This dataset contains Amazon reviews on 4 domains: 
Kitchen items (K), DVD (D), Books (B) and Electronics (E). Instances of the dataset are shown in Fig. \ref{dataset} (c),(d).
The dimensionality of the bag-of-word features was reduced by keeping 
the top 400 features having maximum mutual information with class labels. 
This pre-processing was also carried out in~\citep{sun2016return,gong2013connecting} 
without losing performance. 
For each domain, we used 1000 positive and 1000 negative reviews. 
For each domain-adaptation task, 
we used 1600 samples (800 positive and 800 negative) 
from each domain as the training dataset. 
The remaining 400 samples (200 positive and 200 negative) were used for testing. 
The classifier used is a 1-NN classifier since it is parameter free. 
The mean-accuracy was reported over 10 random training/test splits.

We compared our proposed approach to a recently 
proposed unsupervised domain-adaptation approach known 
as Correlation Alignment (CORAL)~\citep{sun2016return}. 
CORAL is a simple and efficient approach that aligns the input feature distributions 
of the source and target domains by exploring their second-order statistics. 
Firstly, it computes the covariance statistics in each domain 
and then applies whitening and re-coloring linear transformation to the source features. 
Results in Table~\ref{senti} showed that our proposed method outperforms 
CORAL in all the domain-adaptation tasks. 
Our proposed method has better performance because CORAL matches covariances 
while our method matches samples explicitly through point-wise and pair-wise matching. 
Moreover, CORAL does not use source-domain label information. 
Our method uses source-domain label information 
through the group-lasso regularization. 
However, CORAL is quite fast in transforming 
the source samples compared to our method. 
For a single trial, CORAL took about a second while our proposed method 
took about a few minutes.
\begin{table}[h]
\centering
\caption{{Accuracy results of unsupervised domain-adaptation tasks for the Amazon reviews dataset.}}
\label{sentitab}
\begin{tabular}{c c c c c c c }
\hline
\textbf{Tasks} & \textbf{K$\rightarrow$D} & \textbf{D$\rightarrow$B} & \textbf{B$\rightarrow$E} & \textbf{E$\rightarrow$K} & \textbf{K$\rightarrow$B} & 
\textbf{D$\rightarrow$E}\\ \hline
NA    & 58.6                        & 63.4                     & 58.5                    & 66.5            & 59.3        & 57.9\\ 
CORAL & 59.9                        & 66.5                     & 59.5                    & 67.5                    & 59.2 & 59.5\\
Ours  & \textbf{63.5}               & \textbf{69.5}            & \textbf{62.0}           & \textbf{69.5}   & \textbf{64.5}  & \textbf{61.2}     \\ 
Ours (RV)  & {60.9}               & \textbf{69.5}            & \textbf{62.0}           & \textbf{69.5}   & \textbf{64.5}  & {59.0}     \\\hline
\end{tabular}
\label{senti}
\end{table}

\section{{Limitations}}
In this paper, we have assumed that the dimensionality of the source and target feature space is the same. Our approach cannot be directly used in cases when the dimensionality of the features are not the same. However, we can think of two ways in which this problem can be solved in conjunction with our approach. Firstly, we can add a preprocessing step where source and target domain is mapped to a latent space of the same dimensionality. Secondly, we can think of modifying the first-order matching term from $|| \bC\bX^t-\bX^s||^2_\mathcal{F}$ to $|| \bC\bX^t-\bX^s\bW||^2_\mathcal{F}$. $\bW$ belongs to $\mathbb{R}^{d \times d'}$ where $\bW$ maps from the source feature space $\mathbb{R}^{d}$ to the target feature space $\mathbb{R}^{d'}$. But it would require properly regularizing $\bW$. 

Another question regarding our approach is whether our method is applicable to structured data. Structured data is stored in the form of databases or tables. These kind of data might contain numerical, categorical or date-time variables. The main problem in using structured data for our domain adaptation method is the presence of categorical variables. The presence of these discrete attributes would make the problem discontinuous and would not allow optimization to converge. However, we can use entity embedding \citep{guo2016entity}, a recently developed method to map these categorical variables to an Euclidean Space. We can then use the embeddings of the categorical data as features for domain adaptation. However, we are yet to have standard domain adaptation datasets for structured data to test upon.

\section{Conclusions}
In this paper, 
we have described a correspondence-mapping method for unsupervised domain adaptation, 
which matched samples in the source domain with the samples in the target domain. 
Our proposed method is inspired from image registration, 
which alternately finds correspondences between samples 
and mapping between the corresponding samples. 
We proposed a convex-optimization-based approach to find 
the correspondence that consists of three cost terms: 
one for point-to-point similarity (first order), 
one for local structural similarity (second order), 
and another for class-based regularization. 
We have averted memory-efficiency problems of our optimization procedure 
by using the conditional gradient approach. 
We further averted the time-efficiency problem by solving 
the linear programming subproblems in conditional gradient method 
using a network simplex method of a min-cost flow problem, 
rather than a general purpose linear programming (LP) solver. 
An experiment on time-efficiency suggests that the network simplex method 
out-performs the general purpose LP solver by a large amount. 
Classification results on datasets of the textual and visual domain suggested 
that our proposed method outperformed other moment-matching methods 
and was comparable to previous sample-matching methods.

We believe that we can further improve
our proposed method in terms of time and accuracy. 
Till now we have taken all the data-samples in the optimization procedure. 
We could efficiently search for ``important'' samples or exemplars that are a small fraction of all the data-samples. 
Consequently, the number of variables to optimize would be less 
and the optimization will be faster. {As a result, the total time of finding the number of exemplars and the domain adaptation optimization procedure will be less.}
Also our method is a non-deep-learning method that directly works on features. 
We feel that extension of our method to deep architectures in terms of 
jointly learning a representation and the correspondences 
and mapping would improve performance in accuracy.

\section*{Acknowledgement}
This work was supported in part by the National
Science Foundation under Grant IIS-1813935.
Any opinion, findings, and conclusions or recommendations expressed in this material are
those of the authors and do not necessarily reflect the views of the National Science Foundation.
\appendix
\section*{Appendix}
\subsection*{Proof of convexity of optimization objective function}
Let's prove the convexity of each cost term in Eq.~\eqref{prob1}.
\begin{enumerate}
\item 
The \textit{first-order similarity term} in the objective function is of the form $\|\bA\|_F^2$, 
where $\bA=\mathbf{CX}^t-\bX^s$ is a matrix that linearly depends on $\bC$. 
Here $\bA\mapsto \|\bA\|_F$ is a convex function due to the properties of norm. 
This convex function is composed with the function $x\mapsto x^2$, 
which is increasing and convex on the positive domain $[0,\infty)$. 
Thus, $\bA\mapsto \|\bA\|_F^2$ is the composition of a convex function 
with a convex increasing function, which makes it convex as well.
\item 
The argument for proving the convexity of \textit{second-order similarity term} 
is similar to that of proving convexity of 
first-order similarity term except that $\bA=\mathbf{CD}^t-\bD^s\bC$,
which is linearly dependent on $\bC$.
\item 
Proving the convexity of group-lasso regularization is easier. 
The \textit{group-lasso regularization term} is a summation of  $\ell_2$ norm terms. 
Now the set $\bC$ is a convex set because it follows positivity 
and affine equality constraints~\citep{boyd2004convex}. 
So a subset of variables of $\bC$ will also form a convex set. 
$\ell_2$ norm on any arbitrary such convex subset 
will produce a convex function. 
Summation of convex functions will yield a convex function 
and therefore the group-lasso regularization is convex.
\end{enumerate}
\subsection*{{Derivation of Gradients of the objective function}}
\begin{align*} 
f_1 &=||\bC\bX^t-\bX^s||^2_\mathcal{F}=\text{Tr}((\bC\bX^t-\bX^s)^T(\bC\bX^t-\bX^s)) \\
& =\text{Tr}((\bX^t)^T\bC^{T}\bC\bX^t - (\bX^t)^T\bC^{T}\bX^s -
(\bX^s)^T\bC\bX^t + (\bX^s)^T\bX^{s}) ,
\end{align*}
and its gradient is
\begin{align*} 
\nabla_{\mathbf{C}}f_1& = \frac{\partial \text{{Tr}}((\bX^t)^T\bC^{T}\bC\bX^t)}
{\partial \bC}-\frac{\partial \text{Tr}((\bX^t)^T\bC^{T}\bX^s)}{\partial \bC}
- \frac{\partial \text{Tr}((\bX^s)^T\bC\bX^t)}{\partial \bC} \\
& =2\bC\bX^{t}(\bX^t)^T-\bX^{s}(\bX^t)^T 
- \bX^{s}(\bX^t)^T=2(\bC\bX^{t}-\bX^{s})(\bX^t)^T.
\end{align*} 
Let $r=\frac{n_t}{n_s}$, then
\begin{align*} 
f_2&=||\bC\bD^t-r\bD^s\bC||^2_\mathcal{F}=\text{Tr}((\bC\bD^t-r\bD^s\bC)^T(\bC\bD^t-r\bD^s\bC))\\
&=\text{Tr}((\bD^t)^T\bC^{T}\bC\bD^t - r(\bD^t)^T\bC^{T}\bD^{s}\bC - r\bC^{T}(\bD^s)^T\bC\bD^{t}  + r^2\bC^{T}(\bD^s)^T\bD^{s}\bC) ,
\end{align*}
and its gradient can be obtained as
\begin{align*} 
\nabla_{\mathbf{C}}f_2 &=\frac{\partial \text{Tr}((\bD^t)^T\bC^{T}\bC\bD^t)}{\partial \bC}-
\frac{\partial \text{Tr}(r(\bD^t)^T\bC^{T}\bD^{s}\bC)}{\partial \bC}\\
& \quad - \frac{\partial \text{Tr}(r\bC^{T}(\bD^s)^T\bC\bD^{t})}{\partial \bC} + 
\frac{\partial \text{Tr}(r^2\bC^{T}(\bD^s)^T\bD^{s}\bC)}{\partial \bC}\\
&=2\bC\bD^{t}(\bD^t)^T-
2r\bD^{s}\bC(\bD^t)^T-2r(\bD^s)^T\bC\bD^{t} + 2r^{2}(\bD^s)^T\bD^{s}\bC .
\end{align*} 
$\nabla_\mathbf{C}f_3$ can be found by carrying out 
the partial derivative $\frac{\partial f_3}{\partial [\bC]_{ij}}$ 
with respect to each element $[\bC]_{ij}$ of the correspondence matrix.
\begin{equation*}
\frac{\partial f_3}{\partial [\bC]_{ij}}=
\frac{\partial (\sum_j\sum_{c}||[\bC]_{\mathcal{I}_{c}j}||_2)}{\partial [\bC]_{ij}}=
\frac{\partial (||[\bC]_{\mathcal{I}_{c(i)}j}||_2)}{\partial [\bC]_{ij}}.
\end{equation*}
Here, $c(i)$ is the class corresponding to the $i^{th}$ sample in the source domain.
The other summation terms are omitted because they do not depend on $[\bC]_{ij}$. Using the property that partial derivative of an $\ell_2$-norm with respect to an element;
that is, $\frac{\partial (||\bx||_2)}{\partial x_i} = \frac{x_i}{||\bx||_2}$ , we have
\begin{equation*}
\frac{\partial (||[\bC]_{\mathcal{I}_{c}j}||_2)}{\partial [\bC]_{ij}}=\frac{[\bC]_{ij}}{||[\bC]_{\mathcal{I}_{c}(i)j}||_2} .
\end{equation*}
However, the group-lasso regularization term $f_3$ is not differentiable 
if there exists a class $c$ and an index $j$ 
(corresponding to the $j^{\rm th}$ sample of the target domain) 
such that $||[\bC]_{\mathcal{I}_{c}j}||_2=0$. 
In such a case, we set the partial derivative of the corresponding terms to $0$. 
Thus, the gradient of the group-lasso term is found as follows:
\begin{equation*}   
\frac{\partial f_3}{\partial [\bC]_{ij}}= 
\begin{cases}
 \frac{[\bC]_{ij}}{||[\bC]_{\mathcal{I}_{c}(i)j}||_2}~, & \text{if } ||[\bC]_{\mathcal{I}_{c}(i)j}||_2 \neq 0 ;\\
    0~,              & \text{otherwise;}
\end{cases}
\end{equation*}
where $c(i)$ is the class corresponding to the $i^{th}$ sample 
in the source domain.





\bibliographystyle{elsarticle-harv}
\bibliography{TLMaster.bib}







\end{document}

%% file: headerDD.tex
\newcommand{\EQ}{\begin{equation}}
\newcommand{\NQ}{\end{equation}}
\newcommand{\ER}{\begin{eqnarray}}
\newcommand{\NR}{\end{eqnarray}}
\newcommand{\ERS}{\begin{eqnarray*}}
\newcommand{\NRS}{\end{eqnarray*}}
\newcommand{\bit}{\begin{itemize}}
\newcommand{\ben}{\begin{enumerate}}
\newcommand{\eben}{\end{enumerate}}
\newcommand{\ebit}{\end{itemize}}
\newcommand{\bzero}{{\bf 0}}

\newcommand{\bx}{{\bf x}}

\newcommand{\bA}{{\bf A}}

\newcommand{\bC}{{\bf C}}
\newcommand{\bD}{{\bf D}}

\newcommand{\bF}{{\bf F}}
\newcommand{\bG}{{\bf G}}

\newcommand{\bI}{{\bf I}}

\newcommand{\bM}{{\bf M}}

\newcommand{\bP}{{\bf P}}

\newcommand{\bT}{{\bf T}}

\newcommand{\bW}{{\bf W}}
\newcommand{\bX}{{\bf X}}
\newcommand{\bY}{{\bf Y}}





%% file: paperrevision1.bbl
\begin{thebibliography}{53}
\expandafter\ifx\csname natexlab\endcsname\relax\def\natexlab#1{#1}\fi
\expandafter\ifx\csname url\endcsname\relax
  \def\url#1{\texttt{#1}}\fi
\expandafter\ifx\csname urlprefix\endcsname\relax\def\urlprefix{URL }\fi

\bibitem[{Andersen(2013)}]{concomp}
Andersen, E.~D., 2013. Complexity of solving conic quadratic problems.
  http://erlingdandersen.blogspot.com/2013/11/complexity-of-solving-conic-quadratic.html,
  accessed: 2010-09-30.

\bibitem[{Bay et~al.(2006)Bay, Tuytelaars, and Van~Gool}]{surf}
Bay, H., Tuytelaars, T., Van~Gool, L., 2006. Surf: Speeded up robust features.
  In: European Conf. Computer Vision. pp. 404--417.

\bibitem[{Besl and McKay(1992)}]{besl1992method}
Besl, P.~J., McKay, N.~D., 1992. Method for registration of 3-d shapes. In:
  Proc. Intern. Society Optics Photonics. International Society for Optics and
  Photonics, pp. 586--606.

\bibitem[{Blitzer et~al.(2007)Blitzer, Dredze, and
  Pereira}]{blitzer2007biographies}
Blitzer, J., Dredze, M., Pereira, F., 2007. Biographies, bollywood, boom-boxes
  and blenders: Domain adaptation for sentiment classification. In: Proc.
  Annual Meeting Association Computational Linguistics. pp. 440--447.

\bibitem[{Borgwardt et~al.(2006)Borgwardt, Gretton, Rasch, Kriegel,
  Sch{\"o}lkopf, and Smola}]{borgwardt2006integrating}
Borgwardt, K.~M., Gretton, A., Rasch, M.~J., Kriegel, H.-P., Sch{\"o}lkopf, B.,
  Smola, A.~J., 2006. Integrating structured biological data by kernel maximum
  mean discrepancy. Bioinformatics 22~(14), e49--e57.

\bibitem[{Boyd and Vandenberghe(2004)}]{boyd2004convex}
Boyd, S., Vandenberghe, L., 2004. Convex optimization. Cambridge university
  press.

\bibitem[{Bruzzone and Marconcini(2010)}]{bruzzone2010domain}
Bruzzone, L., Marconcini, M., 2010. Domain adaptation problems: A dasvm
  classification technique and a circular validation strategy. {IEEE} Trans.
  Pattern Anal. Mach. Intell. 32~(5), 770--787.

\bibitem[{Chen et~al.(2012)Chen, Xu, Weinberger, and
  Sha}]{chen2012marginalized}
Chen, M., Xu, Z., Weinberger, K., Sha, F., 2012. Marginalized denoising
  autoencoders for domain adaptation. arXiv preprint arXiv:1206.4683.

\bibitem[{Chui and Rangarajan(2003)}]{chui2003new}
Chui, H., Rangarajan, A., 2003. A new point matching algorithm for non-rigid
  registration. Computer Vision and Image Understanding 89~(2), 114--141.

\bibitem[{Courty et~al.(2017)Courty, Flamary, Tuia, and
  Rakotomamonjy}]{courty2016optimal}
Courty, N., Flamary, R., Tuia, D., Rakotomamonjy, A., 2017. Optimal transport
  for domain adaptation. {IEEE} Trans. Pattern Anal. Mach. Intell. 39~(9),
  1853--1865.

\bibitem[{Csurka(2017)}]{csurka2017domain}
Csurka, G., 2017. Domain adaptation for visual applications: A comprehensive
  survey. arXiv preprint arXiv:1702.05374.

\bibitem[{Daum{\'e}~III(2009)}]{daume2009frustratingly}
Daum{\'e}~III, H., 2009. Frustratingly easy domain adaptation. arXiv preprint
  arXiv:0907.1815.

\bibitem[{Donahue et~al.(2014)Donahue, Jia, Vinyals, Hoffman, Zhang, Tzeng, and
  Darrell}]{decaf}
Donahue, J., Jia, Y., Vinyals, O., Hoffman, J., Zhang, N., Tzeng, E., Darrell,
  T., 2014. Decaf: A deep convolutional activation feature for generic visual
  recognition. In: Intern. Conf. Machine Learning. pp. 647--655.

\bibitem[{Duan et~al.(2009)Duan, Tsang, Xu, and Maybank}]{duan2009domain}
Duan, L., Tsang, I.~W., Xu, D., Maybank, S.~J., 2009. Domain transfer svm for
  video concept detection. In: Proc. {IEEE} Conference on Computer Vision and
  Pattern Recognition (CVPR). pp. 1375--1381.

\bibitem[{Duchenne et~al.(2011)Duchenne, Bach, Kweon, and
  Ponce}]{duchenne2011tensor}
Duchenne, O., Bach, F., Kweon, I.-S., Ponce, J., 2011. A tensor-based algorithm
  for high-order graph matching. {IEEE} Trans. Pattern Anal. Mach. Intell.
  33~(12), 2383--2395.

\bibitem[{Farajidavar et~al.(2014)Farajidavar, de~Campos, and
  Kittler}]{farajidavar2014adaptive}
Farajidavar, N., de~Campos, T.~E., Kittler, J., 2014. Adaptive transductive
  transfer machine. In: British Machine Vision Conference.

\bibitem[{Fernando et~al.(2013)Fernando, Habrard, Sebban, and
  Tuytelaars}]{fernando2013unsupervised}
Fernando, B., Habrard, A., Sebban, M., Tuytelaars, T., 2013. Unsupervised
  visual domain adaptation using subspace alignment. In: Proc. {IEEE} Int.
  Conf. Computer Vision. pp. 2960--2967.

\bibitem[{Frank and Wolfe(1956)}]{frank1956algorithm}
Frank, M., Wolfe, P., 1956. An algorithm for quadratic programming. Naval
  Research Logistics (NRL) 3~(1-2), 95--110.

\bibitem[{Ganin et~al.(2016)Ganin, Ustinova, Ajakan, Germain, Larochelle,
  Laviolette, Marchand, and Lempitsky}]{ganin2016domain}
Ganin, Y., Ustinova, E., Ajakan, H., Germain, P., Larochelle, H., Laviolette,
  F., Marchand, M., Lempitsky, V., 2016. Domain-adversarial training of neural
  networks. Journal of Machine Learning Research 17~(59), 1--35.

\bibitem[{Germain et~al.(2013)Germain, Habrard, Laviolette, and
  Morvant}]{germain2013pac}
Germain, P., Habrard, A., Laviolette, F., Morvant, E., 2013. A pac-bayesian
  approach for domain adaptation with specialization to linear classifiers. In:
  Intern. Conf. Machine Learning. pp. 738--746.

\bibitem[{Ghifary et~al.(2015)Ghifary, Bastiaan~Kleijn, Zhang, and
  Balduzzi}]{ghifary2015domain}
Ghifary, M., Bastiaan~Kleijn, W., Zhang, M., Balduzzi, D., 2015. Domain
  generalization for object recognition with multi-task autoencoders. In: Proc.
  {IEEE} Int. Conf. Computer Vision. pp. 2551--2559.

\bibitem[{Gong et~al.(2013)Gong, Grauman, and Sha}]{gong2013connecting}
Gong, B., Grauman, K., Sha, F., 2013. Connecting the dots with landmarks:
  Discriminatively learning domain-invariant features for unsupervised domain
  adaptation. In: Intern. Conf. Machine Learning. pp. 222--230.

\bibitem[{Gong et~al.(2012)Gong, Shi, Sha, and Grauman}]{gong2012geodesic}
Gong, B., Shi, Y., Sha, F., Grauman, K., 2012. Geodesic flow kernel for
  unsupervised domain adaptation. In: Proc. {IEEE} Conference on Computer
  Vision and Pattern Recognition (CVPR). pp. 2066--2073.

\bibitem[{Gopalan et~al.(2011)Gopalan, Li, and Chellappa}]{gopalan2011domain}
Gopalan, R., Li, R., Chellappa, R., 2011. Domain adaptation for object
  recognition: An unsupervised approach. In: Proc. {IEEE} Int. Conf. Computer
  Vision. pp. 999--1006.

\bibitem[{Gopalan et~al.(2014)Gopalan, Li, and
  Chellappa}]{gopalan2014unsupervised}
Gopalan, R., Li, R., Chellappa, R., 2014. Unsupervised adaptation across domain
  shifts by generating intermediate data representations. {IEEE} Trans. Pattern
  Anal. Mach. Intell. 36~(11), 2288--2302.

\bibitem[{Gretton et~al.(2009)Gretton, Smola, Huang, Schmittfull, Borgwardt,
  and Sch{\"o}lkopf}]{gretton2009covariate}
Gretton, A., Smola, A., Huang, J., Schmittfull, M., Borgwardt, K.,
  Sch{\"o}lkopf, B., 2009. Covariate shift by kernel mean matching. Dataset
  Shift in Machine Learning 3~(4), 5.

\bibitem[{Griffin et~al.(2007)Griffin, Holub, and Perona}]{griffin2007caltech}
Griffin, G., Holub, A., Perona, P., 2007. Caltech-256 object category dataset.
  Tech. Rep. 7694, California Institute of Technology.

\bibitem[{Guo and Berkhahn(2016)}]{guo2016entity}
Guo, C., Berkhahn, F., 2016. Entity embeddings of categorical variables. arXiv
  preprint arXiv:1604.06737.

\bibitem[{Huang et~al.(2007)Huang, Smola, Gretton, Borgwardt, and
  Sch{\"o}lkopf}]{huang2007correcting}
Huang, J., Smola, A.~J., Gretton, A., Borgwardt, K.~M., Sch{\"o}lkopf, B.,
  2007. Correcting sample selection bias by unlabeled data. In: Advances in
  Neural Information Processing Systems. p. 601.

\bibitem[{Jaggi(2013)}]{jaggi2013revisiting}
Jaggi, M., 2013. Revisiting frank-wolfe: Projection-free sparse convex
  optimization. In: Intern. Conf. Machine Learning. pp. 427--435.

\bibitem[{Jiang et~al.(2008)Jiang, Zavesky, Chang, and Loui}]{jiang2008cross}
Jiang, W., Zavesky, E., Chang, S.-F., Loui, A., 2008. Cross-domain learning
  methods for high-level visual concept classification. In: Proc. {IEEE} Int.
  Conf. Image Processing. pp. 161--164.

\bibitem[{Kanamori et~al.(2009)Kanamori, Hido, and
  Sugiyama}]{kanamori2009efficient}
Kanamori, T., Hido, S., Sugiyama, M., 2009. Efficient direct density ratio
  estimation for non-stationarity adaptation and outlier detection. In:
  Advances in Neural Information Processing Systems. pp. 809--816.

\bibitem[{Kelly(1991)}]{kelly1991minimum}
Kelly, D.~J., 1991. The minimum cost flow problem and the network simplex
  solution method. Ph.D. thesis.

\bibitem[{Long et~al.(2015)Long, Cao, Wang, and Jordan}]{long2015learning}
Long, M., Cao, Y., Wang, J., Jordan, M.~I., 2015. Learning transferable
  features with deep adaptation networks. In: Intern. Conf. Machine Learning.
  pp. 97--105.

\bibitem[{Long et~al.(2013)Long, Wang, Ding, Sun, and Yu}]{long2013transfer}
Long, M., Wang, J., Ding, G., Sun, J., Yu, P.~S., 2013. Transfer feature
  learning with joint distribution adaptation. In: Proc. {IEEE} Int. Conf.
  Computer Vision. pp. 2200--2207.

\bibitem[{Long et~al.(2016)Long, Wang, and Jordan}]{long2016deep}
Long, M., Wang, J., Jordan, M.~I., 2016. Deep transfer learning with joint
  adaptation networks. arXiv preprint arXiv:1605.06636.

\bibitem[{Maaten and Hinton(2008)}]{maaten2008visualizing}
Maaten, L. v.~d., Hinton, G., 2008. Visualizing data using t-sne. Journal of
  Machine Learning Research 9~(Nov), 2579--2605.

\bibitem[{Mosek(2010)}]{mosek2010mosek}
Mosek, A., 2010. The mosek optimization software. Online at http://www. mosek.
  com 54, 2--1.

\bibitem[{Pan et~al.(2011)Pan, Tsang, Kwok, and Yang}]{pan2011domain}
Pan, S.~J., Tsang, I.~W., Kwok, J.~T., Yang, Q., 2011. Domain adaptation via
  transfer component analysis. IEEE Trans. Neural Networks 22~(2), 199--210.

\bibitem[{Pan and Yang(2010)}]{survey1}
Pan, S.~J., Yang, Q., 2010. A survey on transfer learning. IEEE Trans.
  Knowledge Data Engg. 22~(10), 1345--1359.

\bibitem[{Saenko et~al.(2010)Saenko, Kulis, Fritz, and
  Darrell}]{saenko2010adapting}
Saenko, K., Kulis, B., Fritz, M., Darrell, T., 2010. Adapting visual category
  models to new domains. In: European Conf. Computer Vision. pp. 213--226.

\bibitem[{Si et~al.(2010)Si, Tao, and Geng}]{si2010bregman}
Si, S., Tao, D., Geng, B., 2010. Bregman divergence-based regularization for
  transfer subspace learning. IEEE Trans. Knowledge Data Engg. 22~(7),
  929--942.

\bibitem[{Sugiyama et~al.(2008)Sugiyama, Nakajima, Kashima, Buenau, and
  Kawanabe}]{sugiyama2008direct}
Sugiyama, M., Nakajima, S., Kashima, H., Buenau, P.~V., Kawanabe, M., 2008.
  Direct importance estimation with model selection and its application to
  covariate shift adaptation. In: Advances in Neural Information Processing
  Systems. pp. 1433--1440.

\bibitem[{Sun et~al.(2016)Sun, Feng, and Saenko}]{sun2016return}
Sun, B., Feng, J., Saenko, K., 2016. Return of frustratingly easy domain
  adaptation. In: Thirtieth AAAI Conference on Artificial Intelligence.

\bibitem[{Sun and Saenko(2016)}]{deepCoral}
Sun, B., Saenko, K., 2016. Deep coral: Correlation alignment for deep domain
  adaptation. In: European Conf. Computer Vision Workshops. pp. 443--450.

\bibitem[{Torralba and Efros(2011)}]{torralba2011unbiased}
Torralba, A., Efros, A.~A., 2011. Unbiased look at dataset bias. In: Proc.
  {IEEE} Conference on Computer Vision and Pattern Recognition (CVPR). pp.
  1521--1528.

\bibitem[{Tzeng et~al.(2017)Tzeng, Hoffman, Saenko, and
  Darrell}]{tzeng2017adversarial}
Tzeng, E., Hoffman, J., Saenko, K., Darrell, T., 2017. Adversarial
  discriminative domain adaptation. arXiv preprint arXiv:1702.05464.

\bibitem[{Tzeng et~al.(2014)Tzeng, Hoffman, Zhang, Saenko, and
  Darrell}]{tzeng2014deep}
Tzeng, E., Hoffman, J., Zhang, N., Saenko, K., Darrell, T., 2014. Deep domain
  confusion: Maximizing for domain invariance. arXiv preprint arXiv:1412.3474.

\bibitem[{Weiss et~al.(2016)Weiss, Khoshgoftaar, and Wang}]{survey2}
Weiss, K., Khoshgoftaar, T.~M., Wang, D., 2016. A survey of transfer learning.
  Journal of Big Data 3~(1), 1--40.

\bibitem[{Yang et~al.(2007)Yang, Yan, and Hauptmann}]{yang2007cross}
Yang, J., Yan, R., Hauptmann, A.~G., 2007. Cross-domain video concept detection
  using adaptive svms. In: Proc. {ACM} Int. Conf. on Multimedia. pp. 188--197.

\bibitem[{Zadrozny(2004)}]{zadrozny2004learning}
Zadrozny, B., 2004. Learning and evaluating classifiers under sample selection
  bias. In: Intern. Conf. Machine Learning. p. 114.

\bibitem[{Zheng et~al.(2012)Zheng, Liu, Chellappa, and
  Phillips}]{zheng2012grassmann}
Zheng, J., Liu, M.-Y., Chellappa, R., Phillips, P.~J., 2012. A grassmann
  manifold-based domain adaptation approach. In: Proc. {IEEE} Int. Conf.
  Pattern Recognition. pp. 2095--2099.

\bibitem[{Zhong et~al.(2010)Zhong, Fan, Yang, Verscheure, and
  Ren}]{zhong2010cross}
Zhong, E., Fan, W., Yang, Q., Verscheure, O., Ren, J., 2010. Cross validation
  framework to choose amongst models and datasets for transfer learning. In:
  Joint European Conference on Machine Learning and Knowledge Discovery in
  Databases. pp. 547--562.

\end{thebibliography}
